\documentclass[twoside]{article}

\usepackage[accepted]{aistats2023}
\usepackage[round]{natbib}

\usepackage{hyperref}       
\usepackage{url}            
\usepackage{booktabs}       
\usepackage{amsfonts}       
\usepackage{nicefrac}       
\usepackage{microtype}      
\usepackage{xcolor}         
\usepackage{amsmath,amssymb,amsthm}
\usepackage{comment}
\usepackage{xcolor}
\usepackage{graphicx}
\usepackage{subfigure}
\usepackage{enumitem}

\usepackage{xr}
\makeatletter
\newcommand*{\addFileDependency}[1]{
  \typeout{(#1)}
  \@addtofilelist{#1}
  \IfFileExists{#1}{}{\typeout{No file #1.}}
}
\makeatother

\newcommand*{\myexternaldocument}[1]{%
    \externaldocument{#1}%
    \addFileDependency{#1.tex}%
    \addFileDependency{#1.aux}%
}
\myexternaldocument{supp}

\begin{document}

%

%

\twocolumn[

\aistatstitle{Mind the (optimality) Gap: A Gap-Aware Learning Rate Scheduler for Adversarial Nets}

\aistatsauthor{Hussein Hazimeh \And Natalia Ponomareva}

\aistatsaddress{ Google Research \And  Google Research } ]

\begin{abstract}
Adversarial nets have proved to be powerful in various domains including generative modeling (GANs), transfer learning, and fairness. However, successfully training adversarial nets using first-order methods remains a major challenge. Typically, careful choices of the learning rates are needed to maintain the delicate balance between the competing networks. In this paper, we design a novel learning rate scheduler that dynamically adapts the learning rate of the adversary to maintain the right balance. The scheduler is driven by the fact that 
\textsl{the loss of an ideal adversarial net is a constant known a priori}. The scheduler is thus designed to keep the loss of the optimized adversarial net close to that of an ideal network. We run large-scale  experiments to study the effectiveness of the scheduler on two popular applications: GANs for image generation and adversarial nets for domain adaptation. Our experiments indicate that adversarial nets trained with the scheduler are less likely to diverge and require significantly less tuning. For example, on CelebA, a GAN with the scheduler requires only one-tenth of the tuning budget needed without a scheduler. Moreover, the scheduler leads to statistically significant improvements in model quality, reaching up to $27\%$ in Frechet Inception Distance for image generation and $3\%$ in test accuracy for domain adaptation.
\end{abstract}

\section{Introduction}
Adversarial networks have proved successful in generative modeling \citep{goodfellow2014generative}, domain adaptation \citep{ganin2016domain}, fairness \citep{zhang2018mitigating}, privacy \citep{abadi2016learning}, and other domains. Generative Adversarial Nets (GANs) are a foundational example of this class of models \citep{goodfellow2014generative}. Given a finite sample from a target distribution, a GAN aims to generate more samples from that distribution. This is achieved by training two competing networks. A generator $G$ transforms noise samples into the sample space of the target distribution, and a discriminator $D$ attempts to distinguish between the real and generated samples. To generate realistic samples, $G$ is trained to fool $D$. Adversarial nets used in domains other than generative modeling follow the same principle of training two competing networks. 

Training an adversarial net typically requires solving a non-convex, non-concave min-max optimization problem, which is notoriously challenging \citep{razaviyayn2020nonconvex}. In practice,  first-order  methods are commonly used as a heuristic for this problem. One popular choice is Stochastic Gradient Descent Ascent (SGDA), which is an extension of SGD that takes gradient descent and ascent steps over the min and max problems, respectively\footnote{The steps could be simultaneous or  alternating.}. SGDA and its adaptive variants (e.g., based on Adam) are the defacto standard for optimizing adversarial nets \citep{ganin2016domain,RadfordMC15,arjovsky2017wasserstein}. These methods require choosing two base learning rates\footnote{We use the term base learning rate to refer to the base learning rate in adaptive optimizers and to the learning rate of SGDA.}; one for each competing network. However, adversarial nets are very sensitive to the learning rates \citep{LucicKMGB18}, and careful choices are needed to maintain a balance between the competing networks. In practice, the same learning rate is often used for both networks \citep{wang2021generative}, even though decoupled rates can lead to improvements  \citep{heusel2017gans}. The base learning rates typically used in the literature are constant, but could also be decayed during  training. In either case, these rates do not depend on the current state of the network.

In this paper, we argue that a \textsl{dynamic} choice of the base  learning rate that responds to the current state of the adversarial net can significantly enhance training. Specifically, we propose a learning rate scheduler that dynamically changes the base learning rate of existing optimizers (e.g., Adam), based on the current loss of the network. Our scheduler is driven by the following key observation: in many popular formulations, \textsl{the loss of an ideal adversarial net is a constant known a priori}. For example, an ideal GAN is one in which the distributions of the real and generated samples match. Therefore, we can define an \textsl{optimality gap}, which refers to the gap (absolute difference) between the losses of the current and ideal adversarial nets.


Our main hypothesis is that adversarial nets with smaller optimality gaps tend to perform better---we present empirical evidence that verifies this hypothesis on different loss functions and datasets. Motivated by this hypothesis, our proposed scheduler keeps track of the optimality gap. At each optimization step, the scheduler decides whether to increase or decrease the base learning rate of the adversary (e.g., discriminator), in order to keep the optimality gap relatively small. The base learning rate of the competing network (e.g., generator) is kept constant, since controlling the loss of the adversary (through its base rate) effectively modifies that of the competing network\footnote{If the game is zero-sum, an increase in the objective of the adversary will lead to a decrease in the objective of the competing network with an equal magnitude (and vice versa).}.


We demonstrate the effectiveness of the scheduler empirically in two popular use cases: GANs for image generation and Domain Adversarial Neural Nets (DANN) \citep{ganin2016domain} for domain adaptation. We observe that the scheduler significantly reduces the need for tuning (by $\sim 10$x in many cases) and can lead to statistically significant improvements in the main performance metrics (image quality or accuracy) on five benchmark datasets.



\textbf{Contributions: } 
\textbf{(i)} We present statistical evidence showing that GANs with smaller optimality gaps tend to generate higher quality samples (see Sec.  \ref{sec:ideal_loss}). 
\textbf{(ii)} Motivated by the latter evidence, we propose a novel scheduler that adapts the base learning rate of the adversary to keep the optimality gap relatively small and  maintain a balance with the competing network (see Sec. \ref{sec:scheduler}). \textbf{(iii)} \textcolor{black}{We carry out a large-scale statistical study on GANs and DANN to compare the performance of the  scheduler with popular alternatives. Specifically, we study how the tuning budget and weight initialization affect performance by systematically training over 25,000 GANs. The results indicate that the scheduler can reduce the need for tuning by $\sim$ 10x, improve Frechet Inception Distance in GANs by up to $27\%$, and improve accuracy in DANN by up to $ 3\%$ (see Sec. \ref{sec:experiments}). We provide a simple open-source implementation\footnote{\url{https://github.com/google-research/google-research/tree/master/adversarial_nets_lr_scheduler}} of the scheduler that can be used with any existing optimizer.} 
\vspace{-0.1cm}
\subsection{Related Work}
\vspace{-0.2cm}
Gradient-based methods for non-convex, non-concave min-max problems are known to face difficulties during training and may generally fail to achieve even simple notions of stationarity \citep{razaviyayn2020nonconvex}. In the context of GANs, there has been active research on stabilizing training (with different notions of stability). One important line of work introduces new loss functions or formulations that may be more amenable to first-order methods (e.g., via additional smoothness or avoiding vanishing gradients)  \citep{DBLP:conf/icml/LiSZ15,arjovsky2017wasserstein,mao2017least,zhao2017energy,nowozin2016f,gulrajani2017improved}. Another related approach is to augment existing GAN loss functions with regularizers or perform simple modifications to SGDA (which may be interpreted as regularization) to improve stability  \citep{che2019mode,mescheder2017numerics,nagarajan2017gradient,yadav2018stabilizing,mescheder2018training,xu2020understanding}. Improved architectures have also been vital in successfully training GANs, e.g., see  \citet{RadfordMC15,neyshabur2017stabilizing,lee2021vitgan} and the references therein. See also \citet{karras2020training} for improving stability using data augmentation. Fundamental to all the approaches described above is the choice of the (base) learning rates, which effectively controls the balance between the competing networks. The base rates used in the literature are typically fixed, but may also be decayed during training. In either setting, the base rates used do not take into account  the current state of the network. The main novelty of our  scheduler is that it uses the current state of the network (gauged by the optimality gap) when modifying the learning rate.
\section{Adversarial Nets and their Ideal Loss}  \label{sec:ideal_loss}
We start this section by briefly reviewing a few popular variants of GANs and discussing how their ideal loss can be determined a priori. Then, in Section \ref{sec:opt_gap_sample_quality}, we discuss how the quality of generated samples correlates with the optimality gap. Finally, in Section \ref{sec:DANN}, we introduce DANN and discuss how to estimate its ideal loss.
\subsection{Generative Adversarial Nets (GANs)}
First, we introduce some notation. Let $\mathbb{P}_{\text{r}}$ be the real distribution and  $\mathbb{P}_{n}$ be some noise distribution. The generator $G$ is a function that maps samples from  $\mathbb{P}_{n}$ to the sample space of  $\mathbb{P}_{\text{r}}$ (e.g., space of images). We define $\mathbb{P}_{g}$ as the distribution of $\tilde{x} := G(z)$ where $z \sim \mathbb{P}_{n}$, i.e., $\mathbb{P}_{g}$ is distribution of generated samples. The discriminator $D$ is a function that maps samples from $G$ to a real value. 

\textbf{Standard GAN and its Ideal Loss.} The standard GAN introduced by \citet{goodfellow2014generative} can be written as:
\begin{align*}
  \min_{G} \max_{D} ~~ \mathbb{E}_{x \sim \mathbb{P}_{\text{r}}} \log D(x)  + \mathbb{E}_{\tilde{x} \sim \mathbb{P}_{g}} \log \big(1 -  D(\tilde{x}) \big),
\end{align*}
where $D$ in this case outputs a probability. In practice, we have a finite sample from  $\mathbb{P}_{\text{r}}$ so it is replaced by the corresponding empirical distribution. Moreover, the expectation over $\mathbb{P}_{\text{g}}$ is estimated by sampling from the noise distribution.

We say that a GAN is \textsl{ideal} if the generated and real samples follow the same distribution, i.e.,  $\mathbb{P}_g = \mathbb{P}_r$. When the standard GAN is ideal, the objective function becomes:
\begin{align*}
 \max_{D} ~~ \mathbb{E}_{x \sim \mathbb{P}_{\text{r}}} \Big [ \log D(x) + \log \big(1 -  D(x) \big) \Big].
\end{align*}
The solution to the problem above is given by $D(x) = 0.5$ for all $x$ in the support of $\mathbb{P}_{\text{r}}$. Thus, the optimal objective is $-\log(4)$. Throughout the paper, we will be focusing on the loss, i.e., the negative of the utility discussed above. We will denote the optimal loss of $D$ in an ideal GAN by $V^{*}$, so in this case $ V^{*} = \log(4)$. This quantity allows for computing the optimality gap, which is essential for the operation of the scheduler.

\begin{table*}[t]
  \centering
  \caption{\small{Popular GAN variations considered in this work. Both the discriminator and generator losses are minimized. The value $V^{*}$ denotes the loss  of the discriminator in an ideal GAN.}}
  \label{tab:gans}
  \renewcommand{\arraystretch}{1.3}
  \scalebox{0.7}{
  \begin{tabular}{lllc}
  \toprule
  \textsc{GAN}
  & \textsc{Discriminator Loss (Minimized)}& \textsc{Generator Loss  (Minimized)} & \textsc{Ideal Discriminator Loss} $V^{*}$ \\\toprule
  \textsc{Standard}     
  & $ - \mathbb{E}_{x \sim \mathbb{P}_{r}}[\log(D(x))] -  \mathbb{E}_{\tilde{x} \sim \mathbb{P}_{g}}[\log(1 - D(\tilde{x}))]$
  & $\mathbb{E}_{\tilde{x} \sim \mathbb{P}_{g}}[\log(1 - D(\tilde{x}))]$ & $ \log(4)$ \\\midrule
  \textsc{NSGAN}  
  & $ - \mathbb{E}_{x \sim \mathbb{P}_{r}}[\log(D(x))] -  \mathbb{E}_{\tilde{x} \sim \mathbb{P}_{g}}[\log(1 - D(\tilde{x}))]$
  & $-\mathbb{E}_{\tilde{x} \sim \mathbb{P}_{g}}[\log(D(\tilde{x}))]$ & $ \log(4)$ \\\midrule
  \textsc{WGAN}
  & $ - \mathbb{E}_{x \sim \mathbb{P}_{r}}[D(x)] +  \mathbb{E}_{\tilde{x} \sim \mathbb{P}_{g}}[D(\tilde{x})]$
  & $ - \mathbb{E}_{\tilde{x} \sim \mathbb{P}_{g}}[D(\tilde{x})]$ & $0$ \\\midrule
  \textsc{LSGAN} 
  & $  \mathbb{E}_{x \sim \mathbb{P}_{r}}[(D(x) - 1)^2]  +   \mathbb{E}_{\tilde{x} \sim \mathbb{P}_{g}}[D(\tilde{x})^2]$
  & $ \mathbb{E}_{\tilde{x} \sim \mathbb{P}_{g}}[(D(\tilde{x} - 1))^2]$ & $0.5$ \\\midrule
\end{tabular}}
\end{table*}

\textbf{Popular GAN Variants.} While the standard GAN is conceptually appealing, the gradients of the generator may  vanish early on during training.  To mitigate this issue, \citet{goodfellow2014generative} proposed the non-saturating GAN (NSGAN), which uses the same objective for $D$, but replaces the objective of $G$ with another that (directly) maximizes the probability of the generated samples being real--see Table 1. Similar to the standard GAN, the optimal discriminator loss of an ideal NSGAN is $V^{*} = \log(4)$.


Many follow-up works have proposed alternative loss functions and divergence measures in attempt to improve the quality of the generated samples, e.g., see \citet{arjovsky2017wasserstein,mao2017least,nowozin2016f,li2017mmd} and \citet{wang2021generative} for a survey. In Table 1, we present the objective functions of two popular GAN formulations: Wasserstein GAN (WGAN) and least-squares GAN (LSGAN)  \citep{arjovsky2017wasserstein,mao2017least}. WGAN uses a similar formulation to the standard GAN but drops the log, and $D$ outputs a logit (not a probability). \citet{arjovsky2017wasserstein} shows that under an optimal k-Lipschitz discriminator, WGAN minimizes the Wasserstein distance between the real and generated distributions.  LSGAN uses squared-error loss as an alternative to cross-entropy, and \citet{mao2017least} motivate this by noting that squared-error loss typically leads to sharper gradients.

Similar to an ideal standard GAN, the optimal discriminator losses of ideal WGAN and LSGAN are known constants--see the last column of Table 1 (these constants are derived by plugging  $\mathbb{P}_g = \mathbb{P}_r$ in the discriminator loss). 

\begin{figure*}[htb]
    \vspace{-0.3cm}
    \centering
    \includegraphics[scale=0.37]{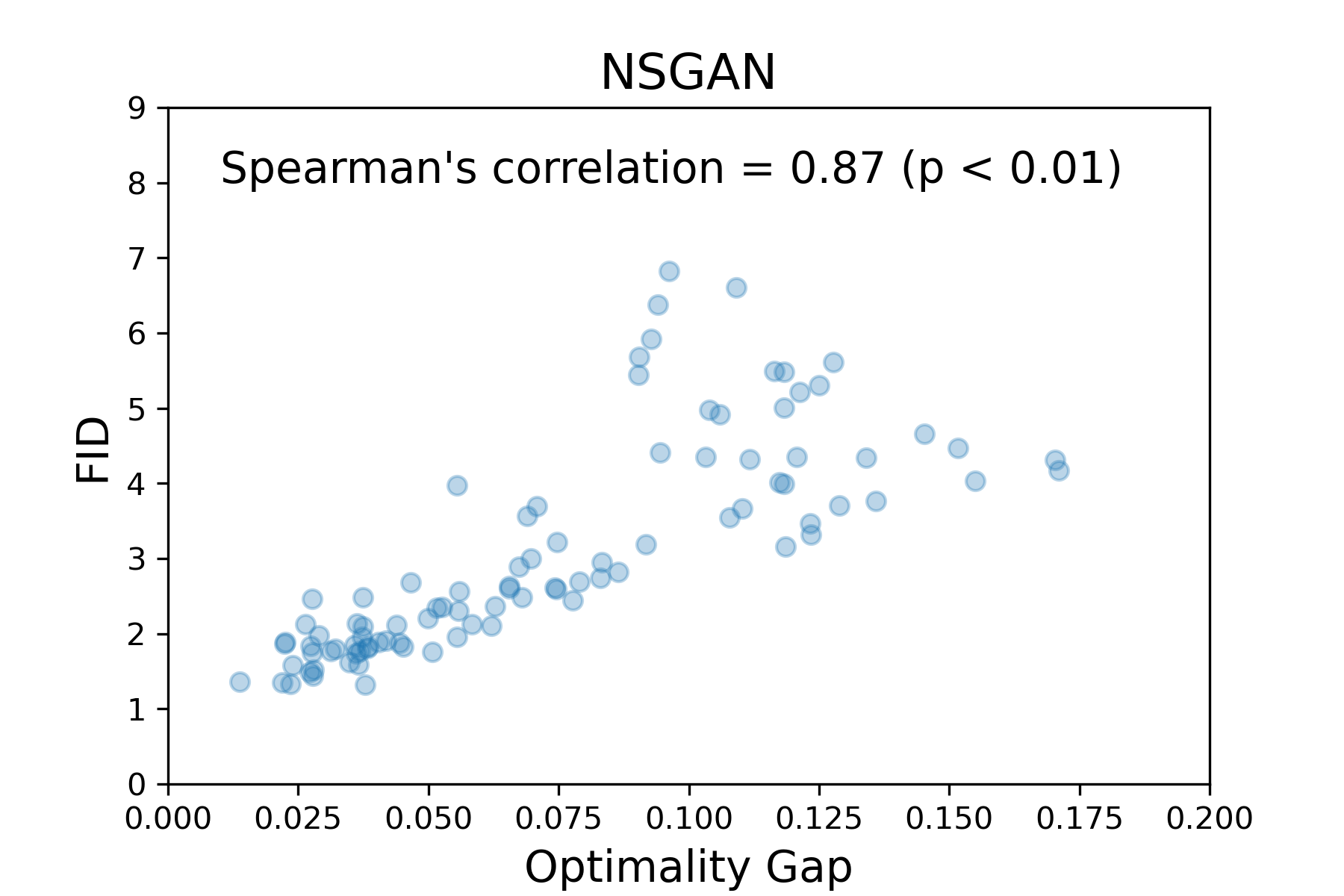}
    \includegraphics[scale=0.37]{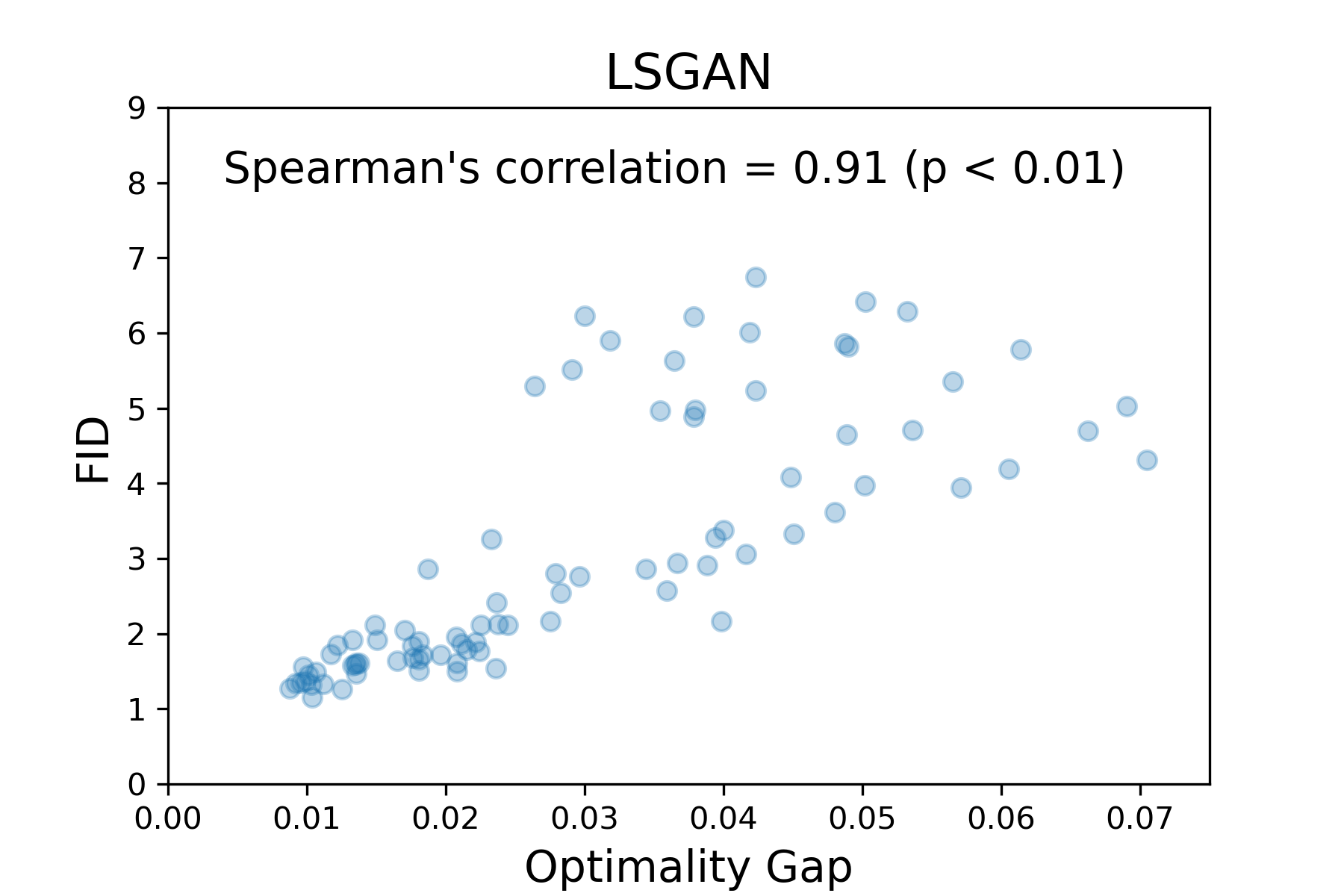}
    \includegraphics[scale=0.37]{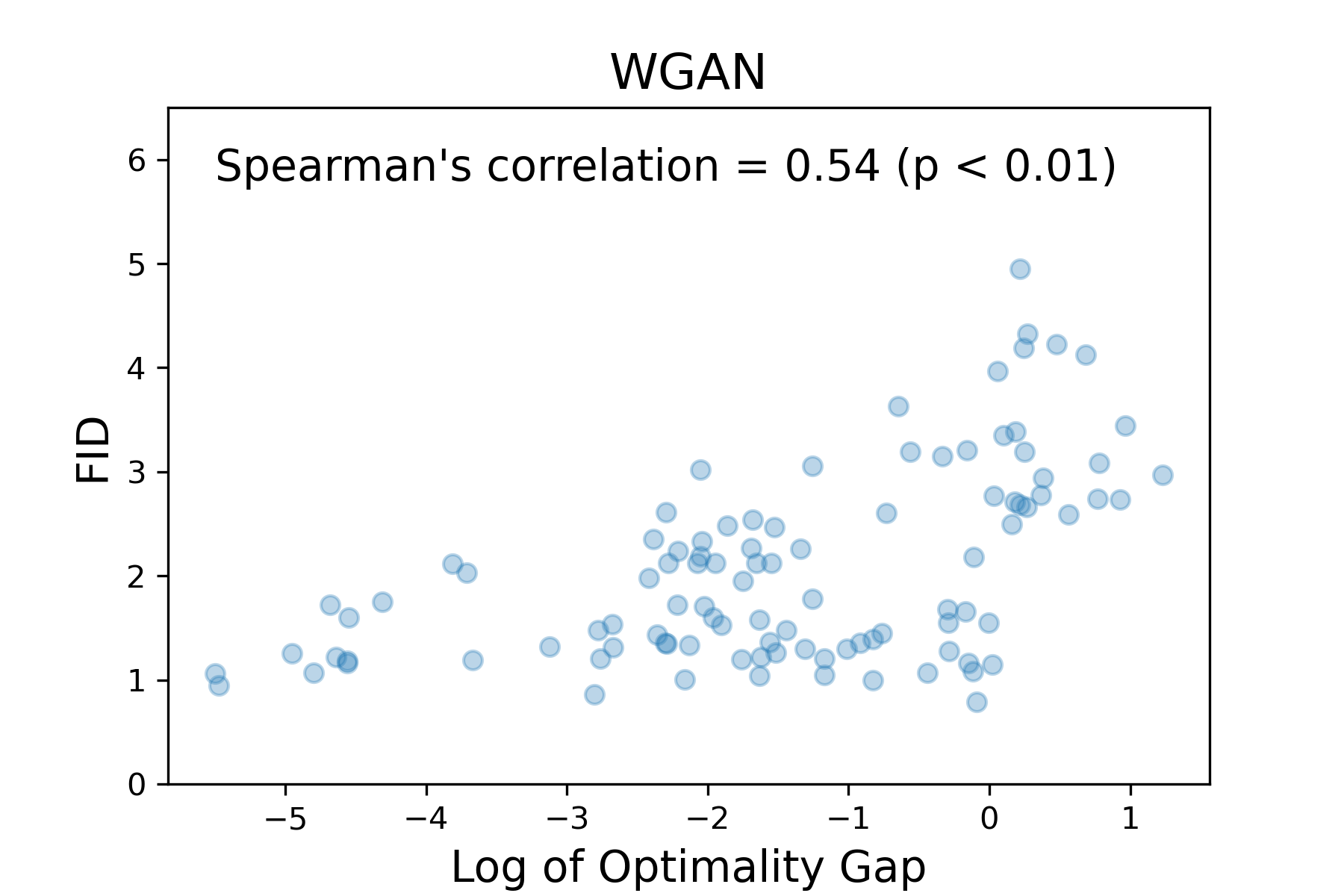}
    \caption{\small Scatter plots of the Frechet Inception Distance (FID) versus the optimality gap on MNIST. Each point corresponds to a particular hyperparameter configuration obtained by random sampling. Lower FID values typically correspond to better image quality. For the three GAN variants, we observe moderate to strong, positive (rank) correlation between FID and the optimality gap. To improve visualization, a small number of outliers was removed--these outliers do not affect correlation and are presented in Appendix \ref{appendix:outlier_viz}. For WGAN, we removed outliers with FID $>$ 5 and applied a log transformation to the gap (since it  varies over $8$ orders of magnitude).}
    \label{fig:corr}
\end{figure*}

\subsubsection{Correlation between the Optimality Gap and Sample Quality} \label{sec:opt_gap_sample_quality}

For all the GAN formulations in Table 1, it is known in theory that if the model capacity is sufficiently high, solving the optimization problem to global optimality leads to an ideal GAN \citep{goodfellow2014generative,arjovsky2017wasserstein,mao2017least}. However, in practice, the capacity of the GAN is limited and optimization is done using first-order methods, which are generally not guaranteed to obtain optimal solutions. Thus, obtaining an ideal GAN in practice is generally infeasible. However, as we demonstrate in our experiments, it is possible to train GANs that are ``close enough'' to an ideal GAN in terms of the loss. Specifically, given a GAN whose discriminator loss is $\hat{V}$, we define the \textsl{optimality gap} as $| \hat{V} - V^{*} |$. Our main hypothesis is:
\begin{center}
\textsl{GANs that achieve smaller optimality gaps tend to generate better samples.}
\end{center}
We stress that this hypothesis applies to GANs that are trained with reasonable hyperparameters and initialization. It is possible to obtain GANs whose optimality gap is 0 or close to 0  without training, e.g., initializing a GAN with  all-zero weights will lead to a 0 gap in standard GAN. 


\textbf{Empirical Evidence.} We validate the hypothesis through multiple experiments on MNIST \citep{DBLP:journals/corr/abs-1708-07747}, Fashion MNIST \citep{DBLP:journals/corr/abs-1708-07747}, CIFAR-10 \citep{krizhevsky2009learning}, and CelebA \citep{liu2015faceattributes}. Next, we briefly discuss one of these experiments and leave the rest to Section \ref{sec:experiments}. We consider generating images from MNIST using a GAN, based on the DCGAN architecture \citep{RadfordMC15}, and we study different GAN variants (NSGAN, LSGAN, and WGAN). We consider $100$ sets of hyperparameter values, drawn randomly, on which we train each GAN (see Section \ref{sec:experiments} for more details). For evaluation, we compute the Frechet Inception Distance (FID) \citep{heusel2017gans}, which is a standard for assessing image quality.

In Figure \ref{fig:corr}, we present scatter plots of FID versus the optimality gap; here each point corresponds to a particular hyperparameter configuration. For the three variants of GAN, we observe medium to strong, positive spearman's correlation between FID and the optimality gap. That is, models with a smaller optimality gap tend to have better image quality. The scheduler we develop (in Sec.  \ref{sec:scheduler}) attempts to keep the optimality gap in check by modifying the learning rate.
\subsection{Domain Adversarial Neural Nets (DANN)} \label{sec:DANN}
DANN is another important example of adversarial nets used in domain adaptation \citep{ganin2016domain}. Given labelled data from a source domain and unlabelled data from a related, target domain, the goal is to train a model that generalizes well on the target. The main principle behind DANN is that for good generalization, the feature representations should be domain-independent \citep{ben2010theory}. DANN consists of: (i) a feature extractor $F$ that receives  features (from either the source or target data) and  generates representations, (ii) a label predictor $Y$ that classifies the source data based on the representations from the feature extractor, (iii) a discriminator $D$--a probabilistic classifier--that takes the feature representations from the extractor and attempts to predict whether the sample came from the source or target domain. Let $\mathbb{P}_{s}$ and $\mathbb{P}_{t}$ be the input distributions of the source and target domains, respectively. At the population level, DANN solves:
\begin{align*}
    \min_{F, Y} \max_{D} ~~  \mathcal{L}_y(F, Y) - \lambda   \mathcal{L}_d(F, D), 
\end{align*}
where $\mathcal{L}_y(F, Y)$ is the risk of the label predictor, $\lambda$ is a non-negative hyperparameter, and $\mathcal{L}_d(F, D)$ is the discriminator risk defined by:
$$- \mathbb{E}_{x \sim \mathbb{P}_{s}} \log \big [ D(F(x)) \big ] -  \mathbb{E}_{\tilde{x} \sim \mathbb{P}_{t}} \log \big [ 1 - D(F(\tilde{x})) \big ].$$
We say that DANN is ideal if the distribution of $F(x), x \sim \mathbb{P}_{s}$ is the same as that of $F(\tilde{x}), \tilde{x} \sim \mathbb{P}_{t}$. By the same reasoning used for standard GAN, the optimal discriminator in this ideal case outputs $0.5$, and thus  $\mathcal{L}_d(F, D) =  \log(4)$. However, generally, $\lambda$ controls the extent to which the two distributions discussed above are matched, and thus the optimal $\mathcal{L}_d(F, D)$ generally depends on $\lambda$. Very small values of $\lambda$ may\footnote{Small values are not always guaranteed to lead to a discriminator that distinguishes well. This depends on a combination of factors including the architecture and the input distributions. As a trivial example, if DANN is supplied with identical domains ($\mathbb{P}_{s} = \mathbb{P}_{t}$), the optimal discriminator outputs $0.5$ for any $\lambda \geq 0$.} lead to a discriminator that distinguishes well between the two domains. On the other hand, by increasing $\lambda$, we can get arbitrarily close the ideal case (where the discriminator outputs $0.5$). In theory, for effective domain transfer, $\lambda$ needs to be chosen large enough so that discriminator is well fooled \citep{ben2010theory}, so for such $\lambda$'s we  expect the optimal $\mathcal{L}_d(F, D)$ to be roughly close to $\log(4)$. Finally, similar to GANs, we remark that the ideal case is typically infeasible to achieve in practice (due to several factors, including using first-order methods and limited capacity); but controlling the optimality gap  can be useful, as we demonstrate in our experiments. 
\section{Gap-Aware Learning Rate Scheduling} \label{sec:scheduler}
In Section \ref{sec:ideal_loss}, we presented empirical evidence that validates our hypothesis that GANs with smaller optimality gaps tend to generate higher quality samples. In this section, we put the hypothesis into action and design a learning rate scheduler that attempts to keep the gap relatively small throughout training. Besides the hypothesized improvement in sample quality, keeping the optimality gap small throughout training can mitigate potential drifts in the loss (e.g., the discriminator loss dropping towards zero), which may lead to more stable training. Next, we describe the optimization setup and then introduce the scheduling algorithm.

\textbf{Optimization Setup.} We assume that the optimization problem of the adversarial net is cast as a minimization over both the loss of the adversary $D$ (e.g., the discriminator in a GAN) and the loss of the competing network $G$ (e.g., the generator in a GAN). 
We focus on the popular strategy of optimizing the two competing networks simultaneously using (minibatch) SGD\footnote{This is SGDA if optimization over $D$ is formulated as maximization.}. We use the notation $\alpha_d$ to refer to the learning rate of $D$. The learning rate scheduler will modify $\alpha_d$ throughout training whereas the learning rate of $G$ remains fixed. We note that the scheduler can be applied to adaptive optimizers (e.g., Adam or RMSProp) as well--in such cases, $\alpha_d$ will refer to the base learning rate. We denote by $V_d$ the current loss of $D$ (a scalar representing the average of the loss over the whole training data). The scheduler takes $V_d$ and D's ideal loss $V^{*}$ as inputs and outputs a scalar, which is used as a multiplier to adjust $\alpha_d$.

\textcolor{black}{\textbf{Effect of D's learning rate  on the optimality gap.}  Recall that in our setup $D$ and $G$ are simultaneously optimized. During each optimizer update, $D$ aims to decrease $V_d$ while $G$ typically aims to increase $V_d$. The optimizer update may increase or decrease $V_d$, depending on how large $D$'s learning rate is w.r.t. that of $G$. If D's learning rate is sufficiently larger, we expect $V_d$ to decrease after the update, and otherwise, we expect $V_d$ to increase. This intuition will be the basis of how the scheduler controls the optimality gap.}

Next, we introduce the scheduling mechanism, where we  differentiate between two cases: \textbf{(i)} $V_d \geq V^{*}$ and \textbf{(ii)} $V_d < V^{*}$.

\textbf{Scheduling when $V_d \geq  V^{*}$.} 
\textcolor{black}{First, we give an abstract definition of the scheduler and then define the scheduling function formally. In this case, the current loss of $D$ is larger than $V^{*}$, so to reduce the gap, we need to decrease $V_d$. As discussed earlier, this effect can be achieved by increasing D's learning rate sufficiently. Therefore, when $V_d \geq  V^{*}$, we design the scheduler to increase the learning rate, and we make the increase proportional to the gap $(V_d - V^{*})$, so that the scheduler focuses more on larger deviations from optimality.}


There are a couple of important constraints that should be taken into account when increasing the learning rate. First, the increase should be bounded because too large of a learning rate will lead to convergence issues. Second, we need to control the rate of increase and ensure the chosen rate works in practice (e.g., too fast of a rate can lead to sharp changes in the loss and cause instabilities). Next, we define a function that satisfies the desired constraints.

We introduce a scheduling function $f: \mathbb{R} \to \mathbb{R}$, which takes $x := (V_d - V^{*})$ as an input and returns a multiplier for the learning rate. That is, the new learning rate of the discriminator (after scheduling) will be $\alpha_d \times f(x)$. To satisfy the  constraints discussed above (boundedness and rate control), we introduce two user-specified parameters: $f_{\text{max}} \in [1, \infty)$ and $x_{\text{max}} \in \mathbb{R}_{> 0}$. The function $f$ interpolates between the points $(0, 1)$ and $(x_{\text{max}}, f_{\text{max}})$ and caps at $f_{\text{max}}$, i.e., $f(x) = f_{\text{max}}$ for $x \geq x_{\text{max}}$. Here $x_{\text{max}}$ is viewed as a parameter that controls the rate of the increase--a larger $x_{\text{max}}$ leads to a slower rate, and thus the scheduler becomes less stringent. There are different possibilities for interpolation. In our experiments, we tried linear and exponential interpolation and found the latter to work slightly better. Thus, we use exponential interpolation and define $f$ as:
\begin{align} \label{eq:f(x)}
    f(x) = \min \Big \{  [f_{\text{max}}]^{x/x_{\text{max}}}, f_{\text{max}}  \Big \}.
\end{align}
Note that since $f_{\text{max}} \geq 1$, we always have $f(x) \geq 1$ for $x \geq 0$, so the learning rate will increase after scheduling. Moreover, the learning rate is not modified when the gap is zero since $f(0) = 1$. 

\textbf{Scheduling when $V_d \leq V^{*}$.} In this case, reducing the gap requires increasing $V_d$. This can be achieved by decreasing the learning rate of $D$. Similar to the previous case, we design the scheduler so that the decrease is proportional to $(V^{*} - V_d)$ (a non-negative  quantity). More formally, we define a scheduling function $h: \mathbb{R} \to \mathbb{R}$, which takes $x := (V^{*} - V_d)$ as an input and returns a multiplier for the learning rate, i.e., the new learning rate is $\alpha_d \times h(x)$. Similar to the previous case, we introduce two user-specified parameters $h_{\text{min}} \in (0,1]$ (the minimum value $h$ can take) and  $x_{\text{min}} \in \mathbb{R}_{> 0}$ to control the decay rate.  We define $h$ as an interpolation between $(0,1)$ and $(x_{\text{min}}, h_{\text{min}})$, which is clipped from below at $h_{\text{min}}$. We use exponential decay interpolation, leading to:
\begin{align} \label{eq:h(x)}
    h(x) = \max \Big \{  [h_{\text{min}}]^{x/x_{\text{min}}}, h_{\text{min}}  \Big \}.
\end{align}
Since $h_{\text{min}} \in [0,1]$, we always have $h(x) \leq 1$ for $x \geq 0$, implying that the learning rate will decrease after scheduling. We summarize the scheduling mechanism in Algorithm 1.
\begin{enumerate}[leftmargin=*]
\item [] \hspace{-0.5cm}\textbf{Algorithm 1: Gap-Aware Scheduling Algorithm}
\item [] \textbf{Inputs}: Current loss $V_d$ and ideal loss $V^{*}$.
\item [] \textbf{Parameters}: $x_{\text{min}}, x_{\text{max}}, h_{\text{min}} \in (0,1],  f_{\text{max}} \in [1, \infty)$.
{\setlength\itemindent{12pt} \item If $V_d \geq V^{*}$, \textsl{increase} D's learning rate by multiplying it with $f(V_d - V^{*})$ -- see \eqref{eq:f(x)}.}
{\setlength\itemindent{12pt} \item If $V_d < V^{*}$, \textsl{decrease} D's learning rate by multiplying it with $h(V^{*} - V_d)$ -- see \eqref{eq:h(x)}.}
\end{enumerate}

\textcolor{black}{In our experiments, we inspect the optimality gap of GANs trained with and without the scheduler. We observe that the scheduler effectively reduces the optimality gap on all datasets and GAN variants, by up to $70$x (see Table \ref{table:fid_inception}). In most cases, we also observe that models with smaller gaps tend to have better sample quality.}

\textbf{Choice of Parameters.} Based on our experiments, we propose setting the same base learning rate for $G$ and $D$ (and tuning over the learning rate, if the computational budget allows). Under this setting, in all of our GAN experiments and across all datasets, we fix the parameters: $h_{\text{min}} = 0.1$, $f_{\text{max}} = 2$, $x_{\text{min}} = x_{\text{max}} =  0.1 V^{*}$ for NSGAN and LSGAN; and $x_{\text{min}} = x_{\text{max}} =  0.1$ for WGAN. These values were only tuned on MNIST for a very limited number of configurations--see Appendix \ref{appendix:param_discussion} for details and intuition. We found these parameters to transfer well to Fashion MNIST, CIFAR-10, and CelebA. In Section \ref{sec:experiments}, we present a sensitivity analysis in which we vary these parameters over multiple orders of magnitude. The results generally indicate that the scheduler is relatively stable around the default values reported above (but setting these parameters to extreme values may cause instabilities).



For the DANN experiments, we use the same fixed parameters as in GANs ($x_{\text{max}} = x_{\text{min}} = 0.1 V^{*}$), but we consider a single tuning parameter: $V^{*}$. As discussed in Section \ref{sec:DANN}, the optimal discriminator loss in DANN depends on $\lambda$, but we expect it to be roughly close to $\log(4)$ (for good choices of $\lambda$). In our experiments we tune over $V^{*} \in [0.5 \log(4), \log(4)]$ and demonstrate that DANN is not sensitive to $V^{*}$, e.g., with only $10$ random search trials for tuning the base learning rate, $V^{*}$, and $\lambda$, optimizing with the scheduler outperforms its no-scheduler counterpart (with the same tuning budget). 

\textbf{Batch-level Scheduling.} We apply Algorithm 1 at the batch level, i.e., the learning rate is modified at each minibatch update. The motivation behind batch-level scheduling is to keep the loss in check after each update. One popular alternative is to schedule at the epoch level. However, if the epoch involves many batches, the loss may drift drastically throughout one or few epochs (an observation that is common in practice). Scheduling at the batch level can mitigate such drifts early on.

\textbf{Estimating the Current Discriminator Loss.} The scheduling algorithm requires access to the discriminator's loss $V_{d}$ at every minibatch update. The loss should be ideally evaluated over all training examples, however, this is typically inefficient. We resort to an exponential moving average to estimate $V_{d}$. Specifically, let $\hat{V}_{d}$ be the current estimate of $V_d$ and denote by $V_{\text{batch}}$ the loss of the current batch (which is available from the forward pass). The moving average update is: $\hat{V}_{d} \gets \alpha  \hat{V}_{d} + (1-\alpha) V_{\text{batch}}$, where $\alpha \in [0,1)$ is a user-specified parameter that controls the decay rate. In all experiments, we fix $\alpha=0.95$ (no tuning was performed) and initialize with $V^{*}$. We also note that if the training loss is evaluated periodically over the whole dataset (e.g., every number of epochs), the moving average can be reinitialized with this value.
\section{Experiments} \label{sec:experiments}
We study the performance of the scheduler on GANs for image generation and DANN. 

\subsection{GANs}
GANs are generally sensitive to weight initialization and hyperparameters  (especially, learning rate) and require sufficiently large tuning budgets to perform well \citep{LucicKMGB18}. Thus, our main goal is to study if the learning rate scheduler can improve stability and reduce the need for tuning.

\textbf{A Statistical Study.}  We perform a systematic study in which we tune GANs under different tuning budgets and repeat experiments over many random seeds. Our study allows for a rigorous understanding of the statistical significance and stability of the results. The study is large-scale as it involves training over 25,000 GANs (for 100s of epochs each) and requires around 6 GPU years on NVIDIA P100. In this respect, we note that a large part of the literature on GANs reports  results on a single random seed and manually tunes hyperparameters (without reporting the tuning budget)--as reported by  \citet{LucicKMGB18}, this may result in misleading conclusions.

\textbf{Competing Methods, Datasets, and Architecture.} We compare with popular mechanisms for choosing the learning rate, including using the same learning rate for $G$ and $D$, decoupled learning rates (tuned independently) \citep{heusel2017gans}, and a classical scheduler that monotonically decays the learning rate. Since our study involves training a large number of GANs (over 25,000), we consider the following standard datasets that allow for reasonable computation time: CelebA, CIFAR, Fashion MNIST, and MNIST. 
We focus on three popular GAN variants: NSGAN, LSGAN, and WGAN, and use a DCGAN architecture \citep{RadfordMC15}--see Appendix \ref{appendix:experimental_details} for details. Our setup (both datasets and architecture) is standard for large-scale tuning studies of GANs, e.g., see \citet{LucicKMGB18}. 

While it would be interesting to consider larger datasets and architectures, we note that performing such a large-scale study may become computationally infeasible. Moreover, we stress that our goal is to understand how the scheduler performs compared to other alternatives, under a clear, fixed tuning budget. Thus, it would be unfair to compare with models in the literature that do not report the tuning budget and the exact tuning procedure.

\begin{figure*}[htbp]
    \centering
    MNIST \\
\hspace*{-1.5cm}    
    \includegraphics[scale=0.39]{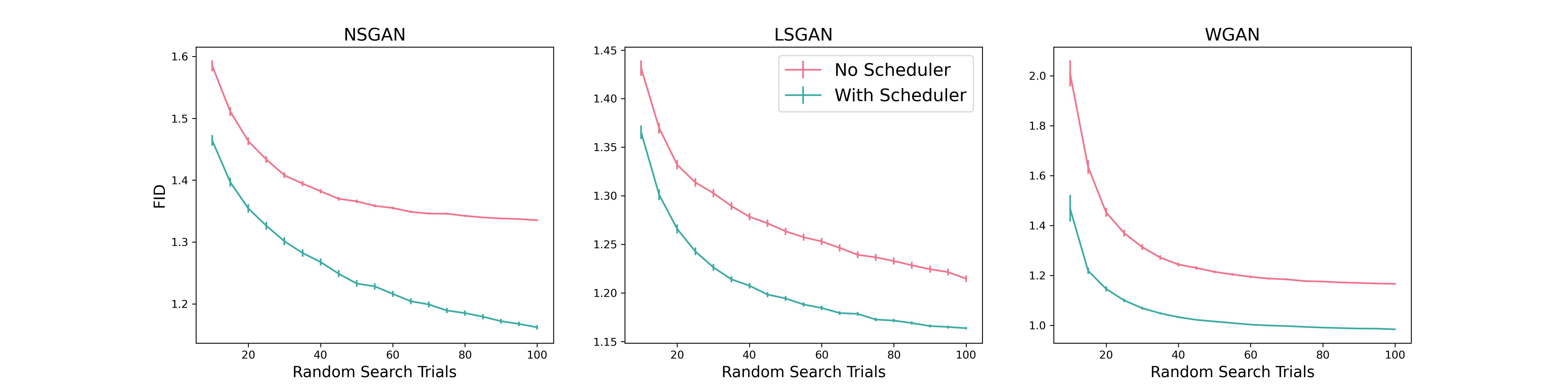} \\[0.0em]
    Fashion MNIST \\
\hspace*{-1.5cm}    
    \includegraphics[scale=0.39]{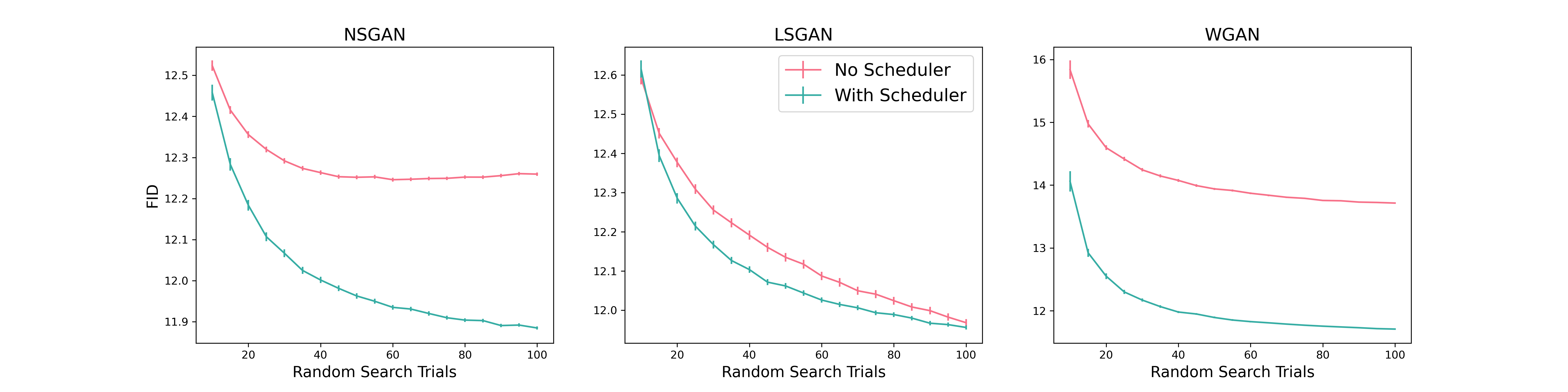} \\[0.0em]
    CIFAR-10 \\
\hspace*{-1.5cm}    
    \includegraphics[scale=0.39]{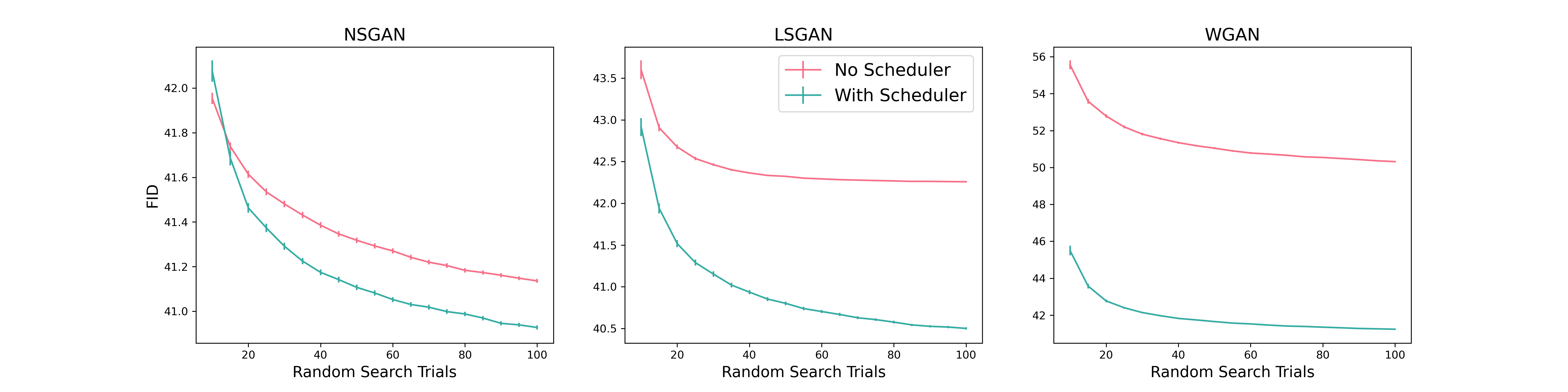} \\[0.0em]
    CelebA \\
\hspace*{-1.5cm}    
    \includegraphics[scale=0.39]{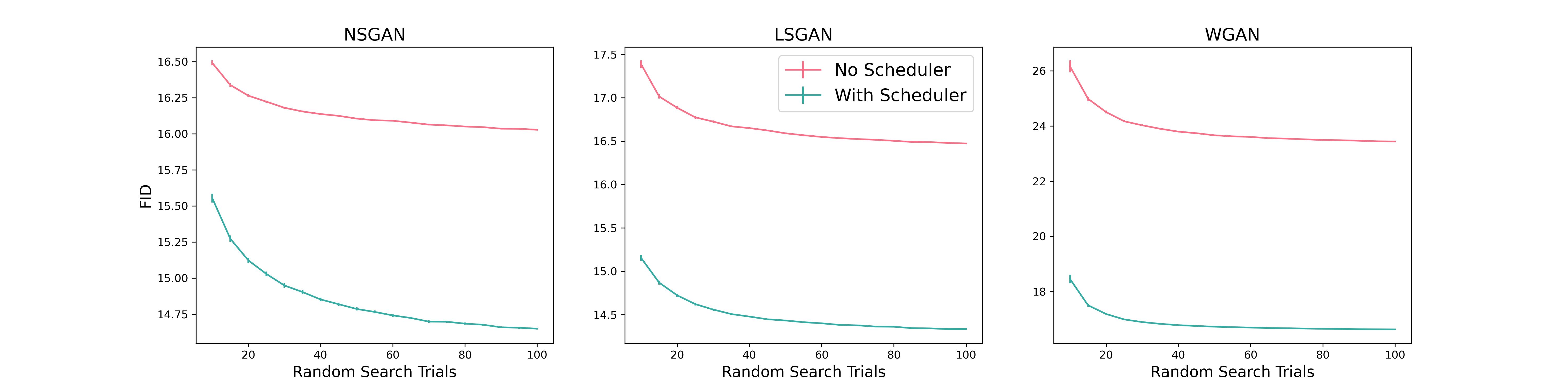}
    \caption{Plots of the best FID as function of the tuning budget. Following \citet{LucicKMGB18}, for each tuning budget $k$, we report the mean and 99$\%$ confidence intervals of the best FID, estimated using 5,000 bootstrap samples of size $k$ from the original $100$ tuning runs.} 
    \label{fig:tune_perf}
\end{figure*}
\begin{table*}[htbp]
\caption{Frechet Inception Distance (lower is better), Inception Score (higher is better), and the Optimality Gap (multiplied by $10^3$) on the test set after tuning. Each entry represents the mean and standard error, computed over 100 training runs (initialized with random seeds). Best values are in bold. An asterisk (*) indicates statistical significance based on a two-sample t-test at a level of $0.01$.}
\label{table:fid_inception}
\centering
\setlength{\tabcolsep}{1.7pt}
\renewcommand{\arraystretch}{1.2}
\scalebox{0.72}{
\begin{tabular}{|l|cccc|cccc|cccc|}
\hline
              & \multicolumn{4}{c|}{FID}         & \multicolumn{4}{c|}{Inception Score} & \multicolumn{4}{c|}{Optimality Gap $\times 10^3$} \\
GAN/Dataset  & MNIST & Fashion & CIFAR & CelebA & MNIST  & Fashion  & CIFAR  & CelebA & MNIST  & Fashion  & CIFAR  & CelebA \\ \hline
NS          & 1.4 (0.03)      &  12.4 (0.04)       &  42.1 (0.09)   & 16.5 (0.06)  & 8.18 (0.01)      &   4.11 (0.01)     &      6.27 (0.01) & 3.1 (0.01)           & 22 (1)      &   34 (1)     &    203 (6) & 513 (8) \\
NS + Sched. & \textbf{1.2} (0.02)*      &  \textbf{12.0} (0.04)*      &    \textbf{41.1} (0.1)*  & \textbf{15.0} (0.07)*   & \textbf{8.23} (0.01)*      &  4.11 (0.01)      &   \textbf{6.41} (0.01)* & \textbf{3.12} (0.0004)        & \textbf{19} (1)      &     34 (1)   &  \textbf{91} (3)*  & \textbf{238} (5)  \\ \hline
LS          & 1.3 (0.02)      & 12.2 (0.04)        &   68.1 (10.28)  & 40.6 (9.5)  & 8.19 (0.01) & 4.04 (0.03)        &    6.01 (0.13)  & 2.97 (0.05)       & \textbf{10} (1)      &     17 (0.4)    &  115 (10)  & 246 (8) \\
LS + Sched. & 1.3 (0.02)      & \textbf{12.1} (0.04)       &    \textbf{40.9} (0.09)*  & \textbf{14.5} (0.05)*  & 8.19 (0.01)      &   \textbf{4.11} (0.01)     &     \textbf{6.4} (0.01)* &  \textbf{3.12} (0.0004)    & 11 (1)      &    17 (0.4)     &  \textbf{27} (0.4)*  & \textbf{111} (3)*     \\ \hline
W           & 1.1 (0.02)      & 13.8 (0.06)        &    49.9 (0.11)   & 23.4 (0.11)   & 8.32 (0.01)      &  4.1 (0.01)       &    5.93 (0.01) & 2.99 (0.01)       & 2143 (169)      &     346 (35)    &     11801 (813)  & 9861 (225) \\
W + Sched.  & \textbf{1.0} (0.02)*      & \textbf{11.6} (0.04)*       &  \textbf{41.8} (0.17)*   & \textbf{17.1} (0.11)* & \textbf{8.4} (0.01)*      & \textbf{4.15} (0.01)*       &      \textbf{6.32} (0.02)* & \textbf{3.1} (0.01)        & \textbf{57} (5)*      &      \textbf{117} (9)*   &   \textbf{825} (96)*   & \textbf{139} (6)* \\ \hline
\end{tabular}}
\end{table*}

\textbf{Experimental Details.} We use Adam \citep{DBLP:journals/corr/KingmaB14} as it is the most popular choice for optimizing GANs \citep{wang2021generative}, and fix the batch size to $256$. On MNIST, Fashion MNIST, and CIFAR, we use  $500$ epochs, and $200$ epochs on CelebA (as it is $\sim 3$x larger than the other datasets). To avoid overfitting, we periodically compute FID on the validation set during training, and upon termination return the version of the model with the best FID (this simulates early stopping). We tune over the following key  hyperparameters: base learning rate(s), $\beta_1$ in Adam, and the clipping weight in WGAN. We consider two settings when tuning the base learning rate: (i) the same rate for both $G$ and $D$, and (ii) two  decoupled rates that are tuned independently \citep{heusel2017gans}. We report the results of (i) in the main paper and the results of (ii) in the appendix--in both cases, the scheduler outperforms its no scheduler counterpart. We use $100$ trials of random search, where in each trial, training is repeated $5$ times over random seeds to reduce variability. We use FID on the validation set as the tuning metric, and we report the final FID results on a separate test set. See Appendix \ref{appendix:experimental_details} for more details.
\subsubsection{Results}
\textbf{Tuning Budget and Performance.} Here we compare the performance with and without the scheduler, using the same base learning rate for $G$ and $D$; see Appendix \ref{appendix:experimental_results} for  decoupled rates \citep{heusel2017gans}. In Figure \ref{fig:tune_perf}, we plot the best FID as a function of the tuning budget. The results indicate that on all datasets, all variants of GANs, and almost every computational budget, the scheduler outperforms the (tuned) optimizer without the scheduler. The improvement reaches up to $27\%$ in some cases, e.g., for WGAN on CelebA. The magnitude of the improvement is more pronounced on CelebA and CIFAR compared to MNIST/Fashion MNIST. This may be attributed to the more complex nature of CelebA and CIFAR, which can require more careful choices of the learning rates. Additionally, we note that the learning rate with the scheduler does not monotonically decrease (as in common learning rate decay)--it varies up and down as the training progresses (see Figure \ref{fig:lr} in the appendix). 
\textbf{Stability.} To get an idea about the stability of the scheduler w.r.t. weight  initialization, we pick the best hyperparameters from the tuning study (after $100$ random search trials), and train each model $100$ times using random seeds. In Table \ref{table:fid_inception}, we report the mean and standard error of both FID and the Inception Score  \citep{salimans2016improved}. The improvements we saw from using the scheduler in the tuning study (represented by Figure \ref{fig:tune_perf}) generalize to this experiment, i.e., the performance of the scheduler does not seem to be sensitive to the random seed. For LSGAN \textsl{without} the scheduler, there are significant outlier runs (the standard error is $\sim 10$) on CIFAR-10 and CelebA--the same observation  was made by \citet{LucicKMGB18} for LSGAN on the latter datasets. In contrast, for LSGAN with the scheduler, we did not observe  outlier runs and this is evidenced by the small standard error ($<0.1$). Thus, for the datasets considered, the scheduler appears to be generally more stable.

\textbf{Optimality Gap.} In Table \ref{table:fid_inception}, we also report the optimality gap of the tuned models (averaged over $100$ randomly initialized training runs). Out of the $12$ dataset/GAN-type pairs, the scheduler achieves a strictly lower optimality gap in $9$ cases, equal gap in $2$ cases, and $1$ worse gap that is statistically insignificant (see LSGAN on MNIST). On CIFAR and CelebA, the scheduler achieves significantly lower gaps, e.g., $70$x lower for WGAN on CelebA. For LSGAN on MNIST and Fashion MNIST, the optimizer without the scheduler already achieves small gaps, so the scheduler does not offer noticeable improvements. Generally, the results are in line with our hypothesis that models with smaller optimality gaps tend to generate better samples.

\begin{figure*}
    \centering
    \includegraphics[scale=0.36]{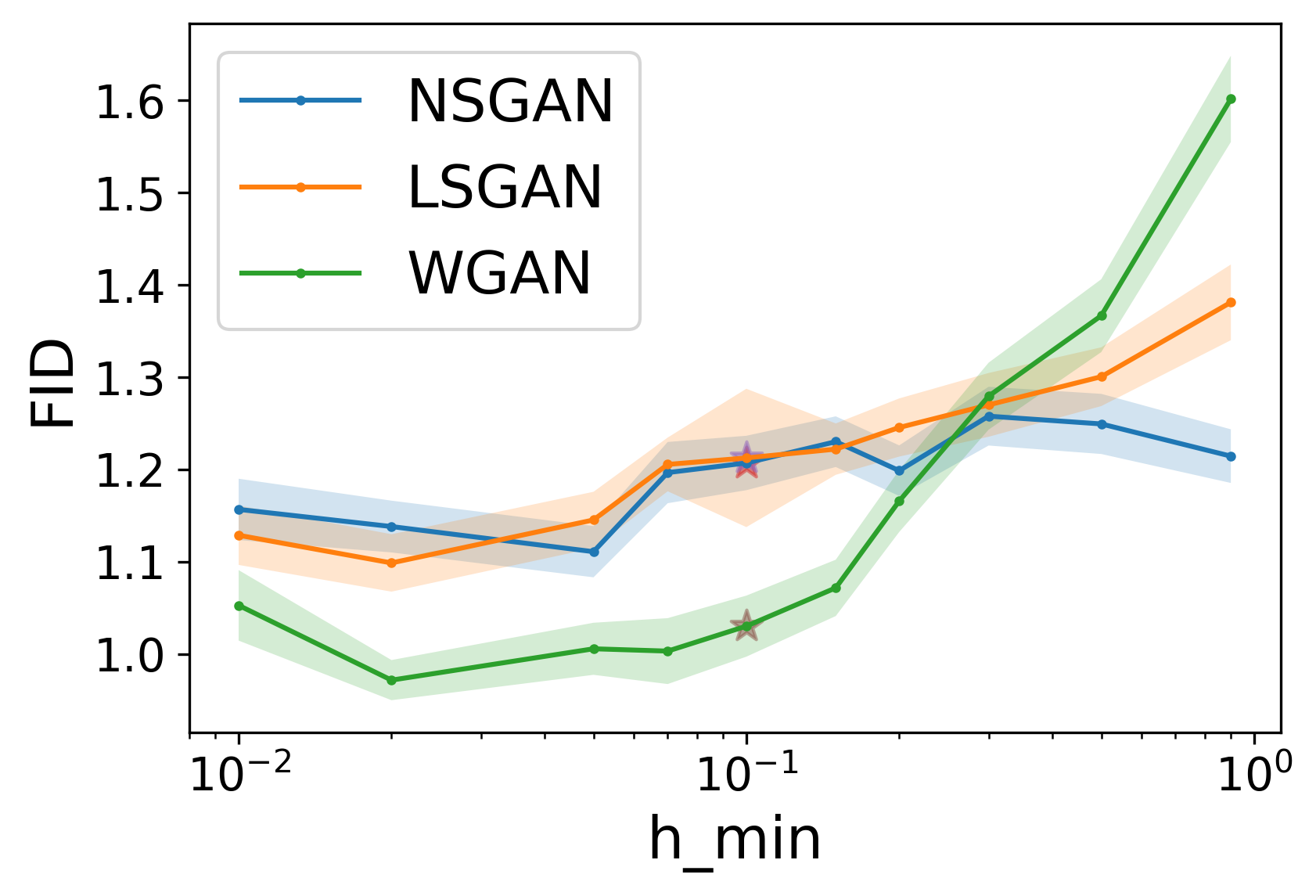}
    \includegraphics[scale=0.36]{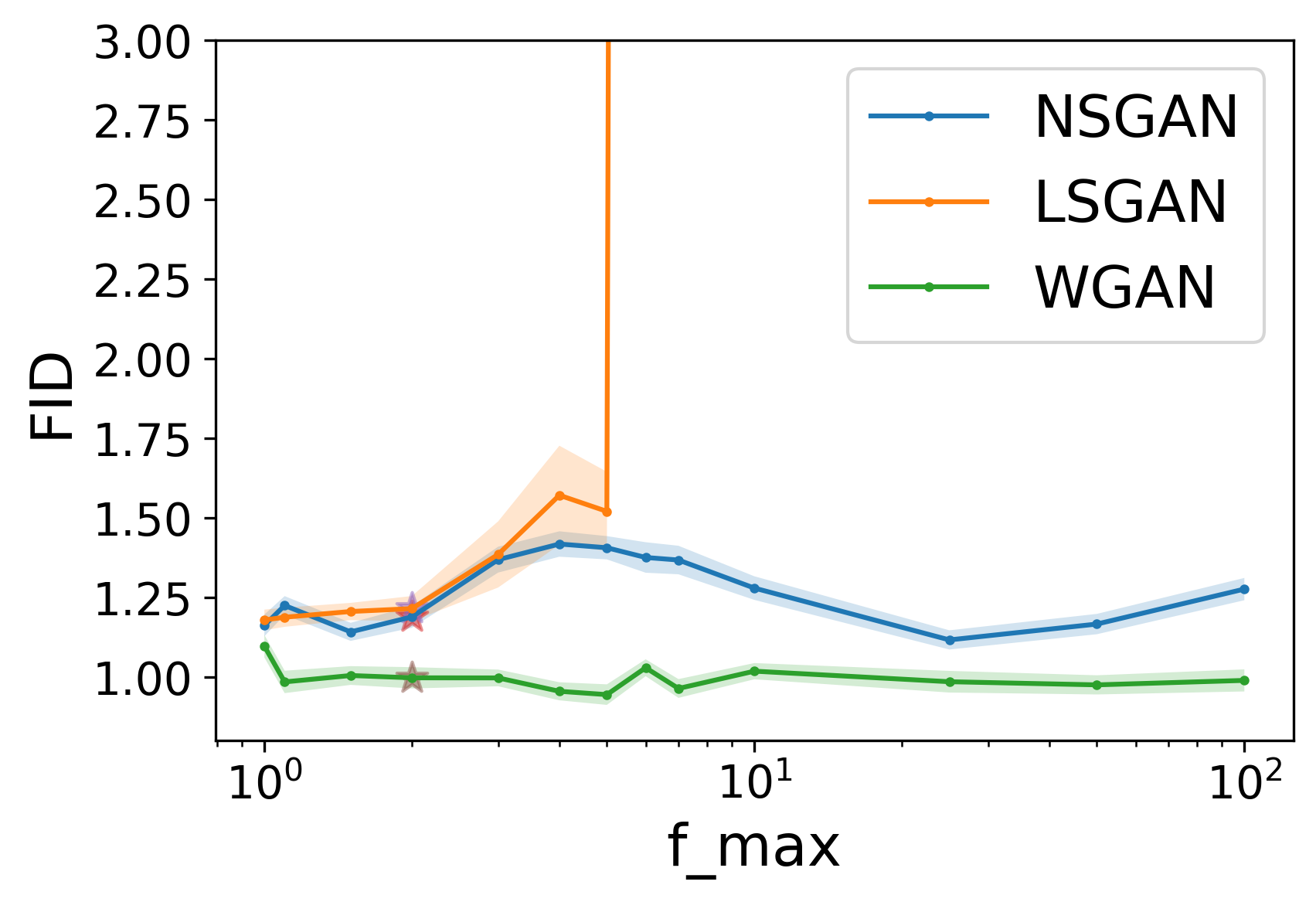}
    \includegraphics[scale=0.36]{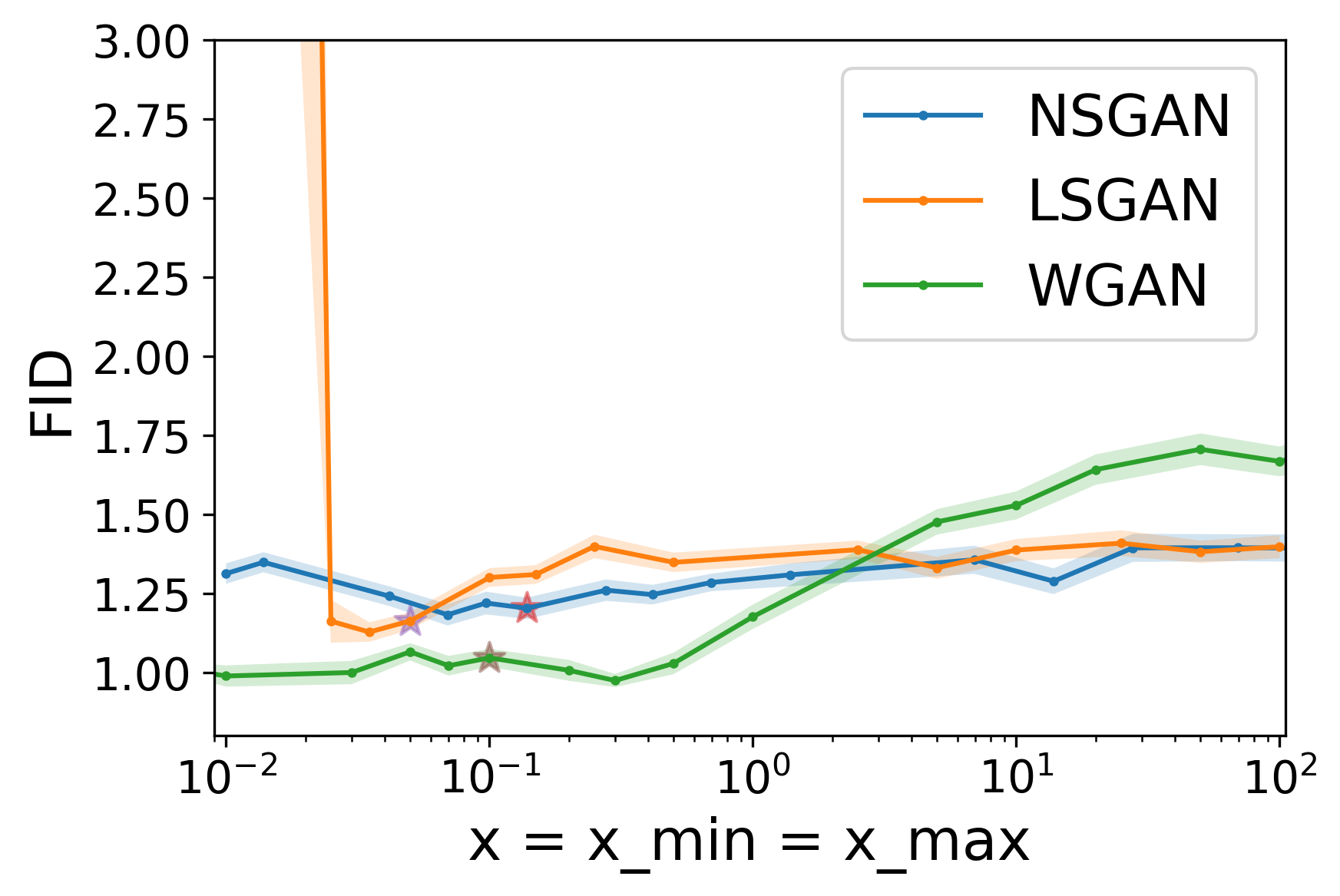}
    \vspace{-0.2cm}
    \caption{{Sensitivity plots for the scheduler applied to NSGAN, LSGAN, and WGAN on MNIST. The x-axis is on a log scale. When varying each parameter, we fix the other parameters to the default values. We repeat training 50 times (using random seeds) and report the mean and standard error (represented by the shaded region). A star represents the default parameter value used in the experiments.}}
    \label{fig:sensitivity}
\end{figure*}

\textbf{Sensitivity Analysis.} We study the sensitivity of the scheduler to its parameters: $h_{\text{min}}$, $f_{\text{max}}$, $x_{\text{min}}$, and  $x_{\text{max}}$. Specifically, we vary the value of each parameter (individually) over multiple orders of magnitude and study the change in FID. When varying a given parameter, we fix the other parameters to their default values (discussed in Section \ref{sec:scheduler}). The analysis is done on MNIST  using the best (tuned) hyperparameters of the GAN, and training is repeated for $50$ random seeds to account for the variability due to initialization. In Figure \ref{fig:sensitivity}, we present sensitivity plots for NSGAN, LSGAN, and WGAN. 

The results indicate that NSGAN and WGAN are relatively insensitive: there is a wide range of values (over one order of magnitude) that lead to good performance, which exceeds that of no scheduler. LSGAN has sharp transitions for large values of $f_{\text{max}}$ (specifically $>$ 5); this is intuitively expected because increasing the learning rate beyond a certain threshold will cause the model to diverge. For very small $x_{\text{min}}$ and $x_{\text{max}}$ ($<$ 0.02), LSGAN performs poorly; this is also expected because such small values force the training loss to be almost constant so essentially the model does not train. We also note that LSGAN is known in the literature to be more sensitive and suffer from frequent failure even for well-tuned hyperparameters, compared to NSGAN and WGAN \citep{LucicKMGB18}.


\textbf{Optimality Gap of G.} Given that the scheduler only controls the learning rate (and loss) of $D$, a natural question is: what happens to the loss of $G$? In Appendix \ref{appendix:experimental_results}, we study the effect of the scheduler on G's loss. Specifically, we measure G's optimality gap, which we define as the absolute difference between G's training and ideal losses. The main conclusion of the experiment is that the scheduler can significantly reduce G's optimality gap, compared to no scheduler.

\textcolor{black}{\textbf{Additional Comparisons.} In Appendix \ref{appendix:experimental_results}, we compare with two additional alternative strategies for choosing the learning rate: (i) decoupled base learning rates  (tuned independently) \citep{heusel2017gans}, and (ii) a classical scheduler that monotonically decreases the learning rate. In both cases, the scheduler reduces the need for tuning (by up to 10x) and significantly improves FID (by up to 38\%). Moreover, we present a comparison between exponential and linear interpolation for the scheduling functions $f(x)$ and $h(x)$.}

\subsection{DANN}
We consider a standard domain adaptation benchmark: MNIST as the source and MNIST-M as the target. MNIST-M consists of MNIST images whose background has been altered  \citep{ganin2016domain}. We conduct a tuning study to understand how DANN with the scheduler compares to (i) DANN without a scheduler and (ii) a model without domain adaptation (i.e., trained only on the source).

\textbf{Experimental Setup.} We use a CNN-based architecture for DANN,  similar to that in \citet{ganin2015unsupervised}, and optimize using SGD with a batch size of $256$. We train for $300$ epochs, computing the validation accuracy at each epoch. At the end of training, we pick the version of the model with the highest validation accuracy (simulating early stopping). Additionally, we tune over the following hyperparameters: learning rate, $\lambda$, and $V^{*}$, using $100$ random search trials, and training is repeated $5$ times per trial (using random seeds). See Appendix \ref{appendix:experimental_details} for details.

\begin{figure}[htbp]
    \centering
    \includegraphics[scale=0.28]{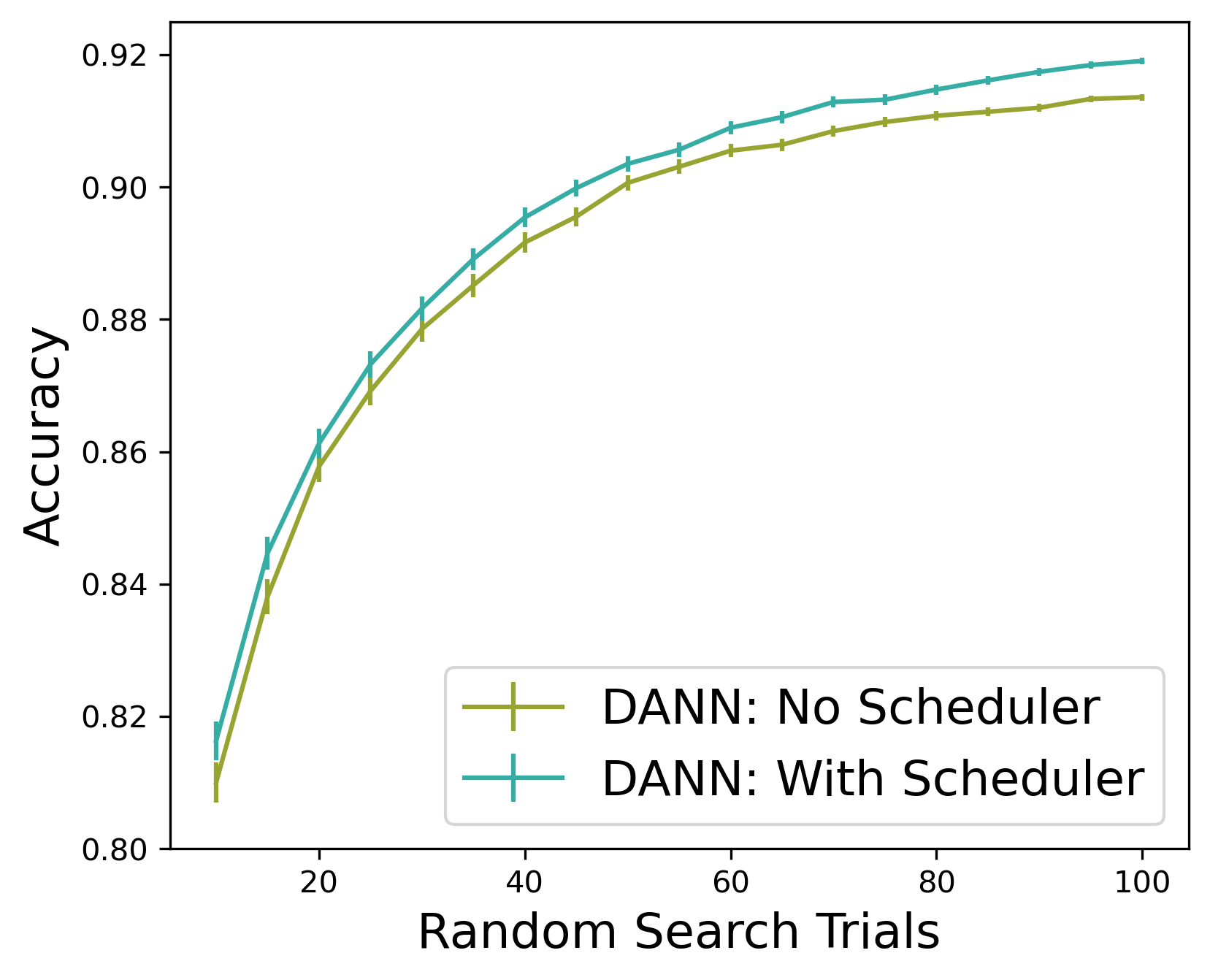}
    \includegraphics[scale=0.29]{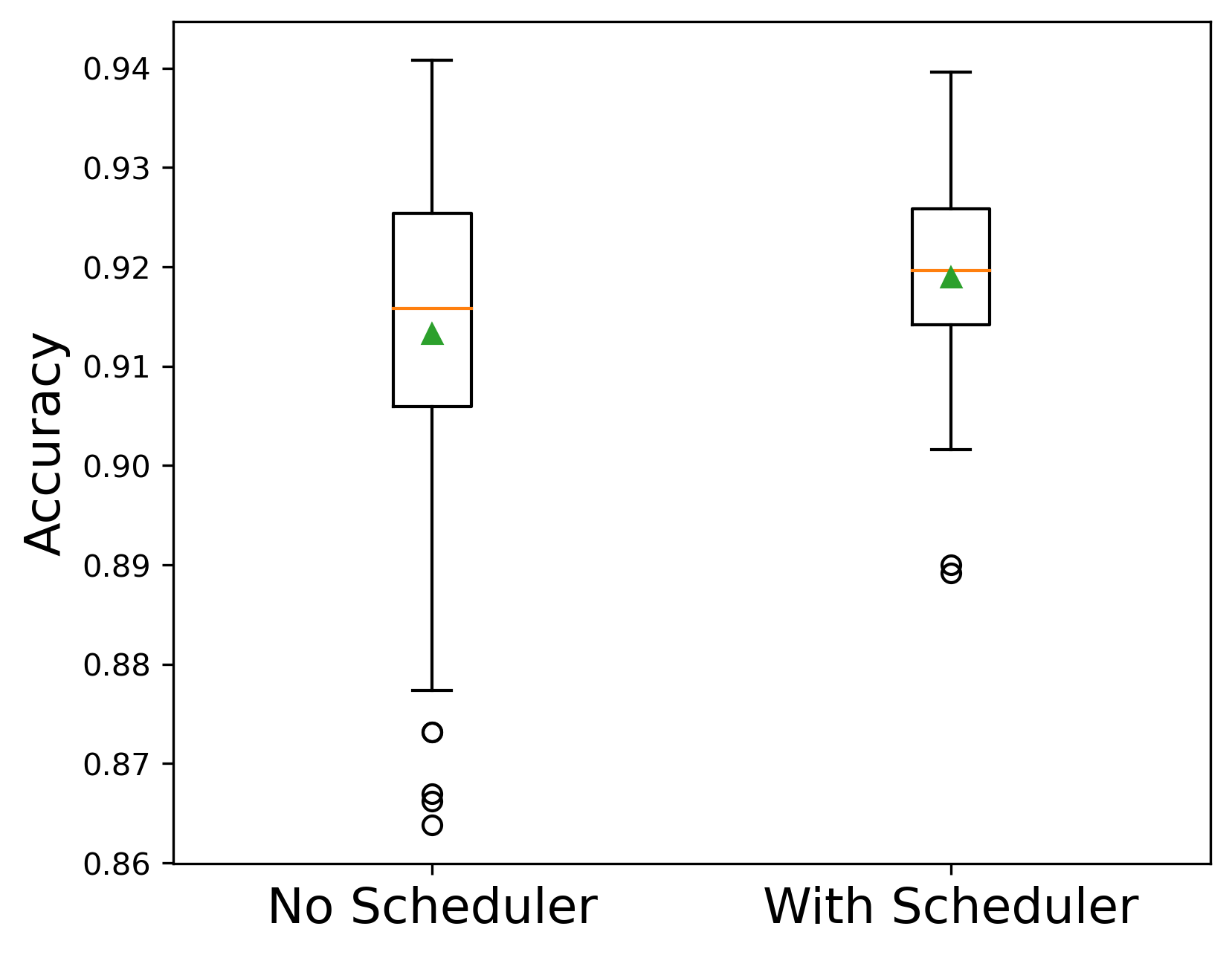}
    \caption{{Domain adaptation (MNIST $\to$ MNIST-M) using DANN. \textbf{[Left]}  Test accuracy of the best model as a function of the tuning budget. The 99\% confidence intervals are estimated using 5000 bootstrap samples. \textbf{[Right]} Training stability: test accuracy of 100 models trained using random initialization and optimal hyperparameters.}}
    \label{fig:DANN}
\end{figure}

\textcolor{black}{\textbf{Results.} In Figure \ref{fig:DANN} (left), we report the test accuracy (on the target) as a function of the tuning budget for DANN with and without the scheduler. The results indicate that the scheduler performs better for every tuning budget. The relative improvement in mean accuracy reaches around  $0.7\%$ at $100$ trials. We also experimented with a source-only model that does not perform domain adaptation (specifically, DANN with $\lambda = 0)$. The accuracy of the source-only model is 60.4\% (with standard error of 0.4\%) at 100 trials, which is significantly lower than the two models in Figure \ref{fig:DANN}. In Figure \ref{fig:DANN} (right), we study the training stability of the model using the optimal hyperparameters (obtained by tuning). Specifically, we report the accuracy of 100 models trained with random initialization. We observe that the scheduler has roughly a 40\% smaller interquartile range, suggesting that it leads to more stable training. Moreover, the scheduler significantly improves the lower tail of the accuracy distribution, e.g., the first quartile and minimum (worst-case) accuracy improve by 1\% and 3\%, respectively.}

\section{Conclusion}
We proposed a novel gap-aware learning rate scheduler for adversarial nets. The scheduler monitors the optimality gap (from an ideal network) during training and modifies the base learning rate to keep the gap in check. This is in contrast to the common choices of base learning rates which do not take into account the gap or the current state of the network. Our experiments on GANs for image generation and DANN for domain adaptation demonstrate that the scheduler can significantly improve performance and reduce the need for tuning.

\clearpage
\bibliography{ref}

\appendix
\renewcommand\thefigure{\thesection.\arabic{figure}}
\setcounter{figure}{0}
\renewcommand\thetable{\thesection.\arabic{table}}
\setcounter{table}{0}
\onecolumn
\section{FID versus Optimality Gap} \label{appendix:outlier_viz}
In Figure \ref{fig:corr}, we removed a small number of outlier points to improve visualization. Below we plot all points including outliers. These outliers have $< 1\%$ effect on Spearman's correlation. For WGAN, a significant number of runs had FID $>$ 5 (corresponding to failures), so we removed these and added additional (non-failing) training runs to have approximately 100 points in the final plot.
\begin{figure}[htbp]
    \centering
    \includegraphics[scale=0.45]{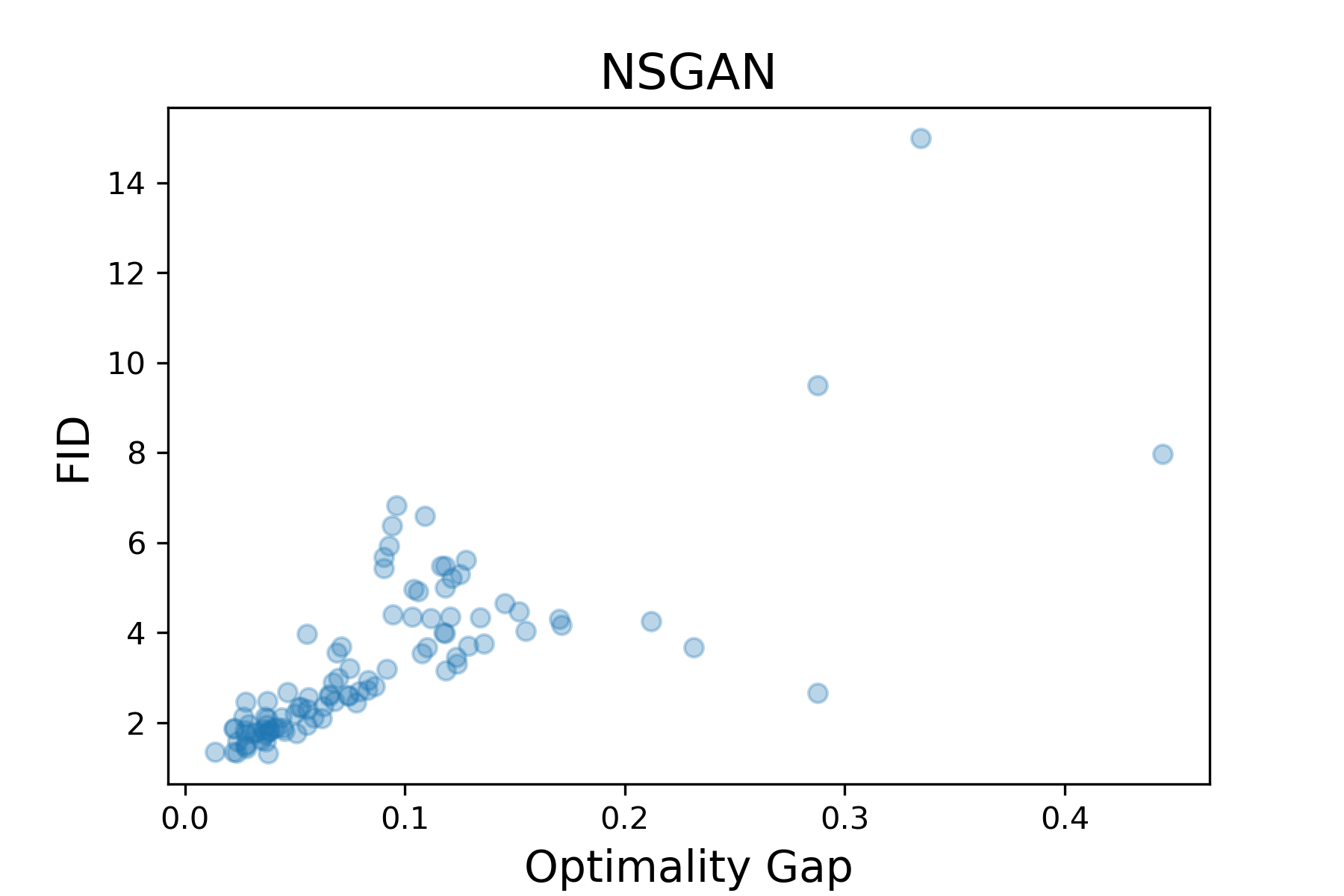}
    \includegraphics[scale=0.45]{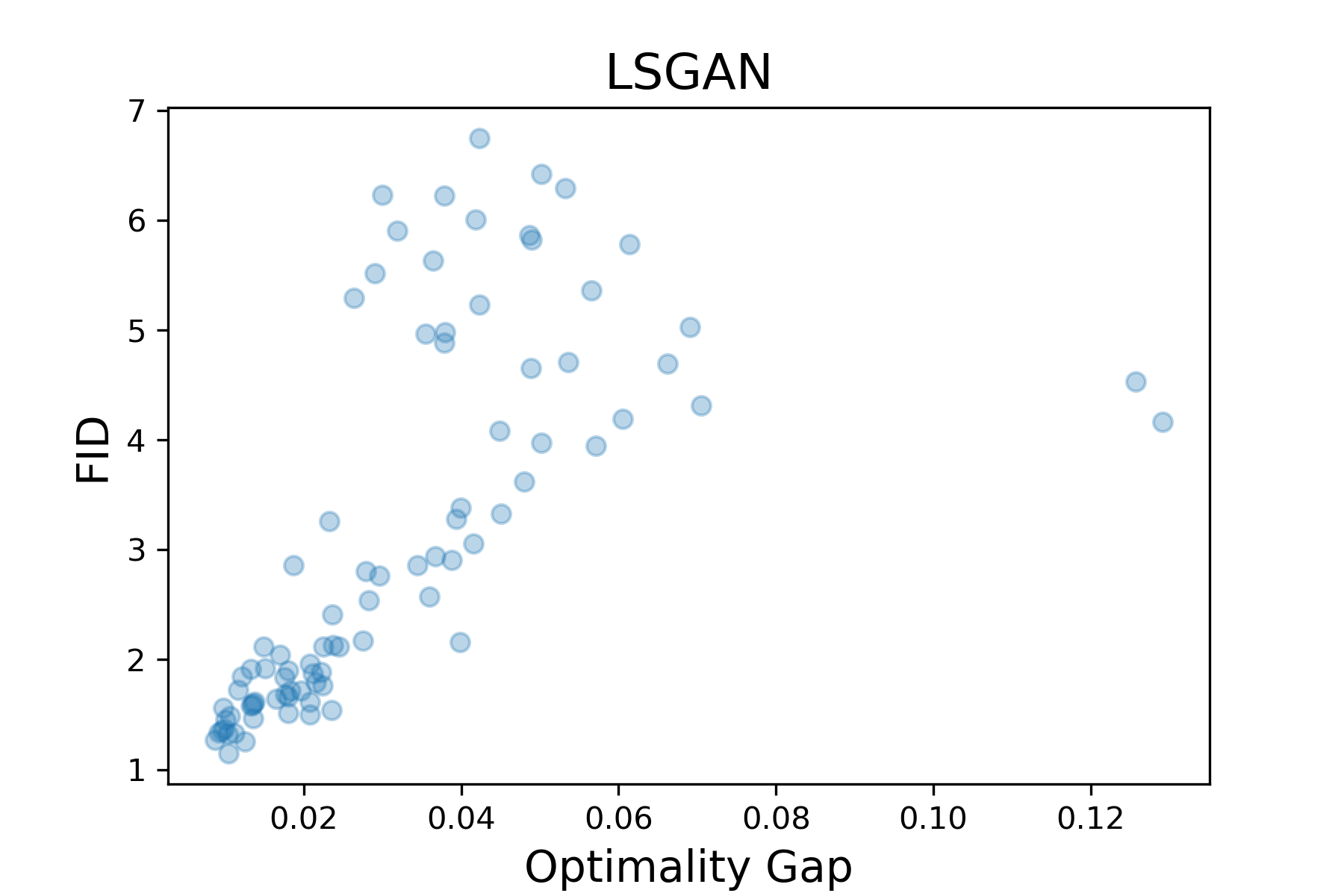}
    \caption{Plots of FID vs. optimality gap for NSGAN and LSGAN with all outliers included.}
\end{figure}

\section{Scheduler's Parameters} \label{appendix:param_discussion}
On MNIST, we tried a total of 8  configurations, which are the Cartesian  product of: $h_{\text{min}} \in \{0.5, 0.1 \}$, $f_{\text{min}} \in \{2, 10 \}$, $x_{\text{min}} = x_{\text{max}} \in \{0.1 V^{*}, 0.5 V^{*} \}$ ($V^{*}$ is dropped for WGAN). We found $h_{\text{min}} = 0.1$, $f_{\text{max}} = 2$, $x_{\text{min}} = x_{\text{max}} =  0.1 V^{*}$ for NSGAN and LSGAN; and $x_{\text{min}} = x_{\text{max}} =  0.1$ for WGAN to work best. We noticed that setting $f_{\text{max}} = 10$ can lead to instabilities--intuitively, there is an upper bound on the learning rate after which the model will diverge. For the opposite direction, i.e., when decreasing the learning rate, having a relatively low floor such as $0.1$ (as opposed to $0.5$) does not seem to cause instabilities (which is expected with small learning rates).

\section{Additional Experimental Results} \label{appendix:experimental_results}

\begin{figure*}[htbp]
    \centering
    \includegraphics[scale=0.35]{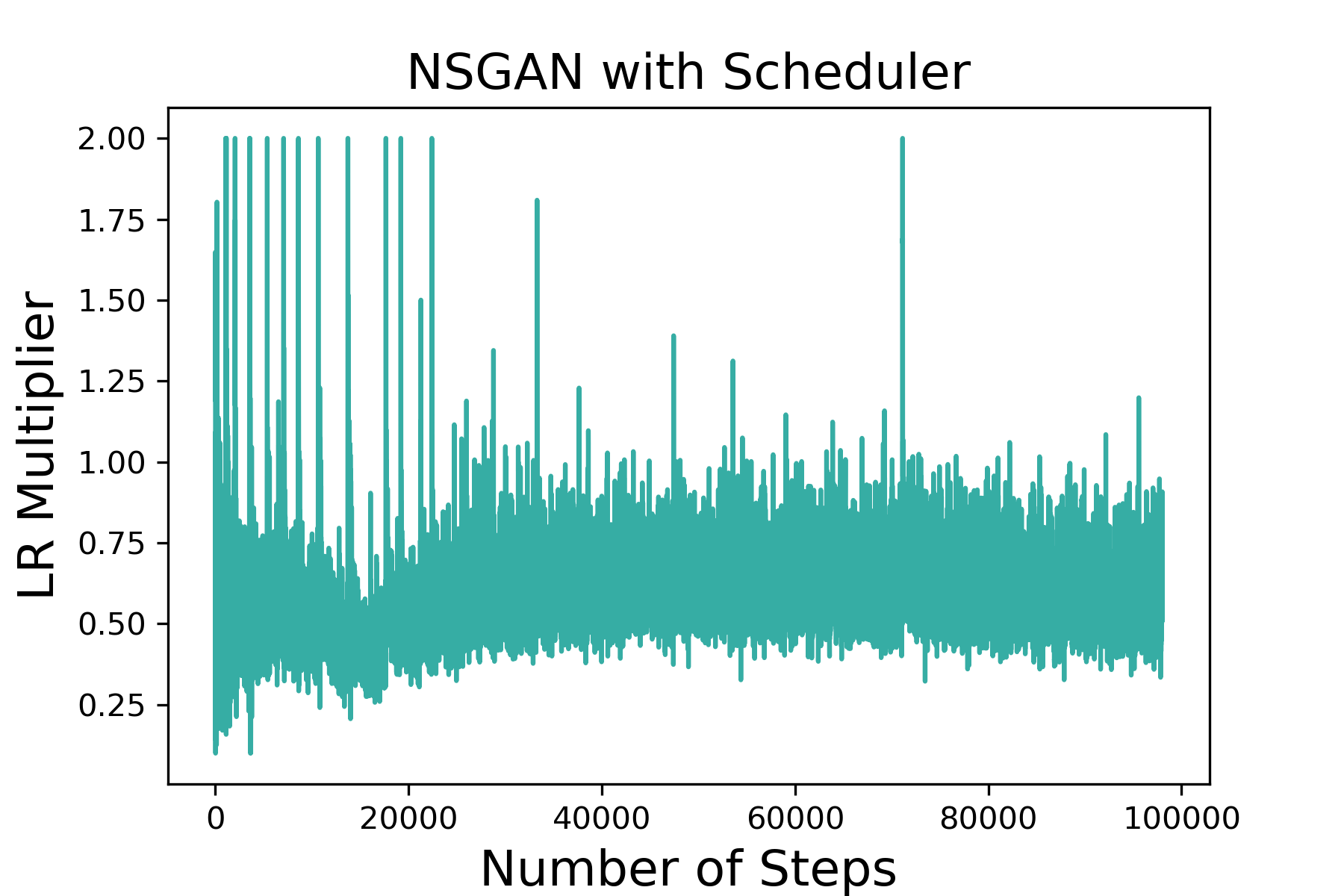} 
    \includegraphics[scale=0.35]{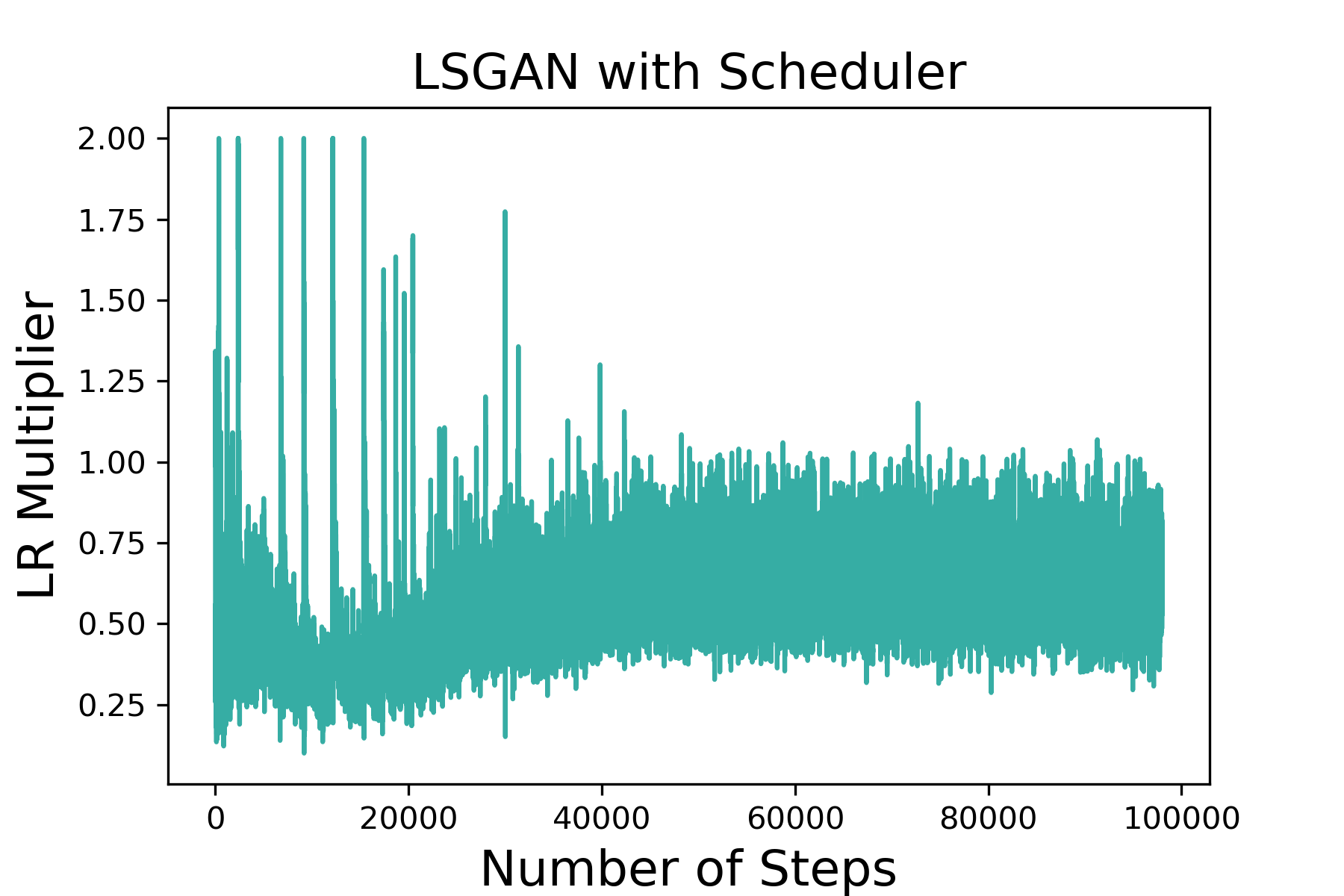} 
    \includegraphics[scale=0.35]{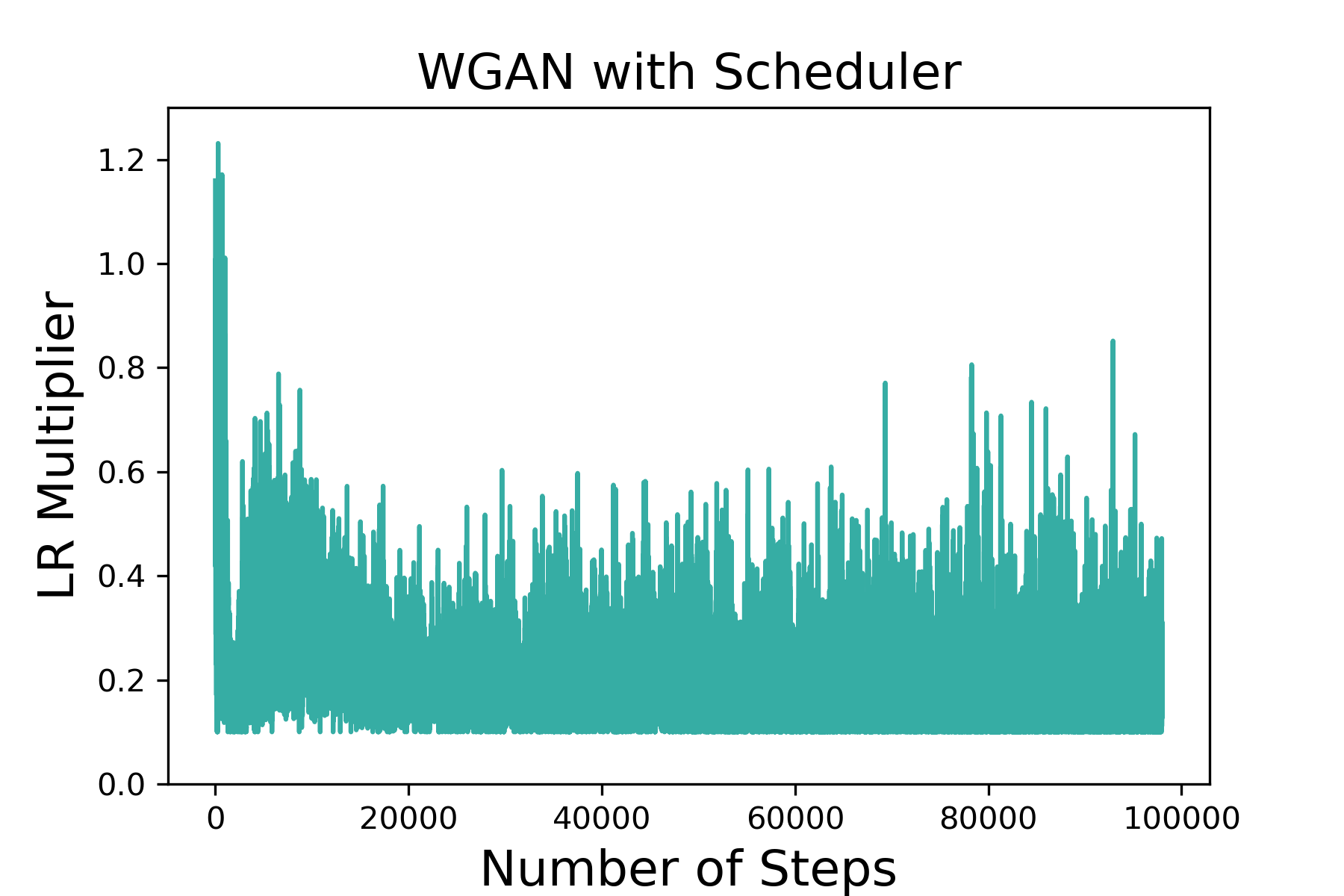} 
    \caption{Change in leaning rate multiplier as the training progresses for NSGAN, LSGAN and WGAN with the scheduler, on MNIST. This demonstrates that the  learning rate varies up and down during training.}
    \label{fig:lr}
\end{figure*}

\textbf{Scheduler's Output.} In Figure \ref{fig:lr}, we visualize the output of the scheduler (the multiplier of the learning rate) for NSGAN, LSGAN, and WGAN using the tuned hyperparameters. The results generally show that the learning rate multiplier continuously goes and up and down during training (based on the current optimality gap). 

\begin{figure*}[htbp]
    \centering
    \small
    MNIST \\
    \includegraphics[scale=0.34]{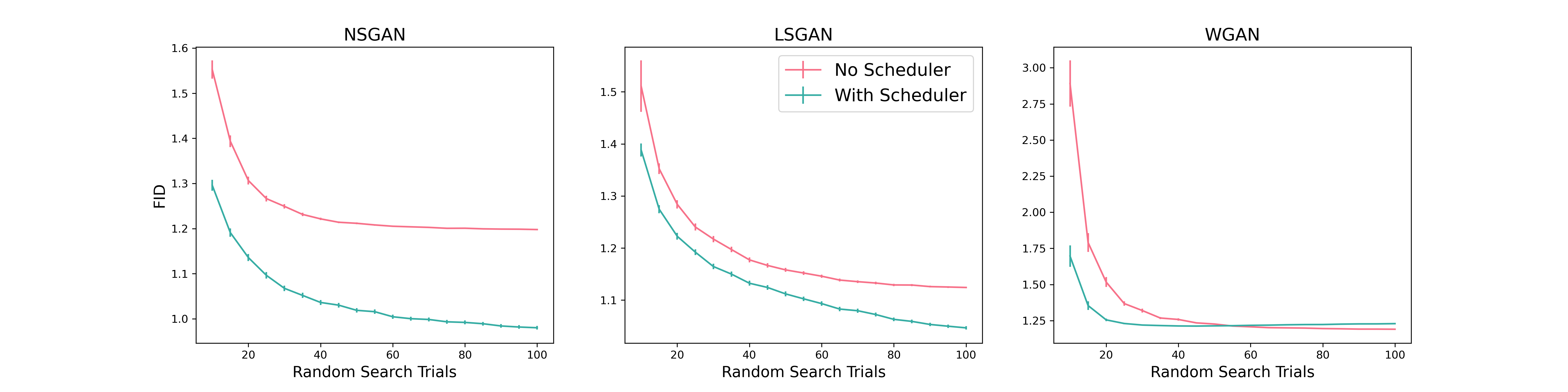} \\[0em]
    Fashion MNIST \\
    \includegraphics[scale=0.34]{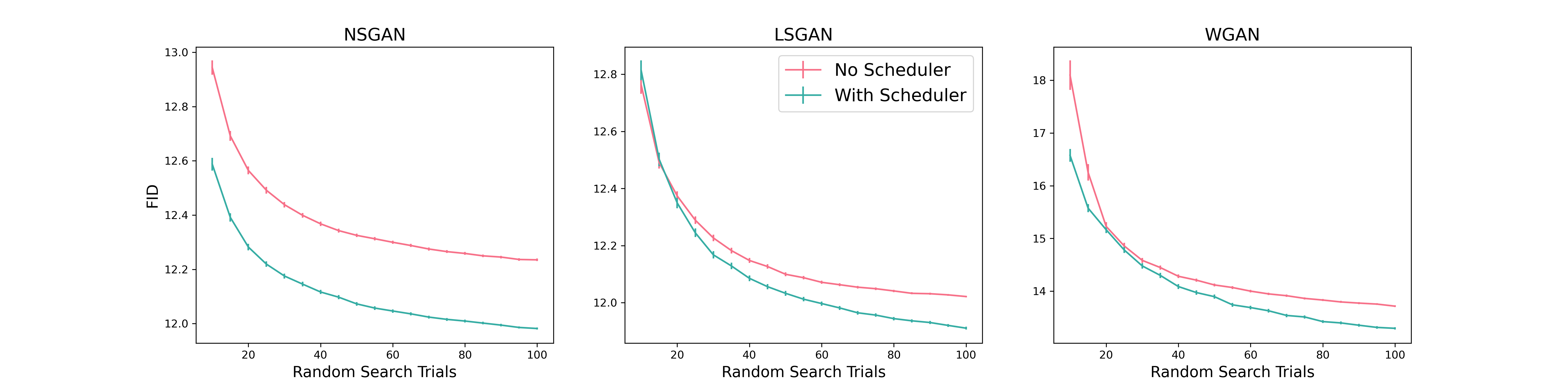} \\[0em]
    CIFAR-10 \\
    \includegraphics[scale=0.34]{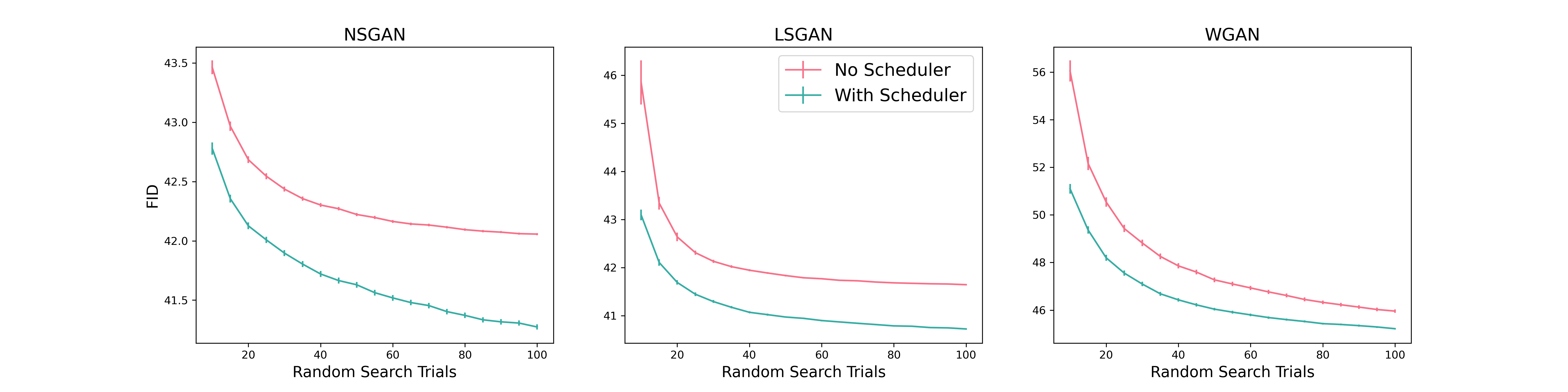} \\[0em]
    CelebA \\ 
    \includegraphics[scale=0.34]{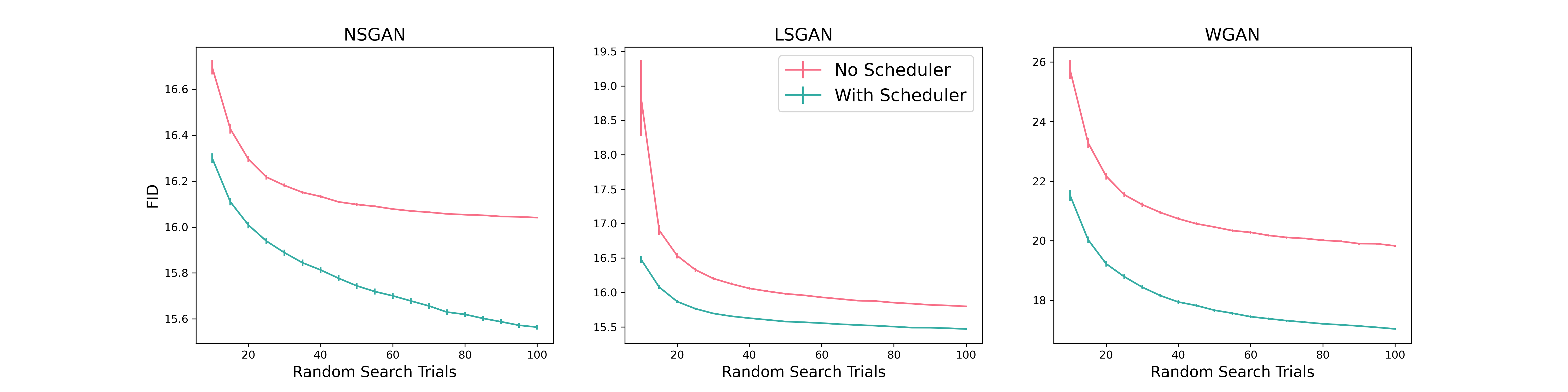}
    \caption{Decoupled Learning Rates:  Plots of the best Frechet Inception Distance (FID) as function of the tuning budget. Following \citet{LucicKMGB18}, for each tuning budget $k$, we report the mean and 99$\%$ confidence intervals of the best FID, estimated using 5,000 bootstrap samples of size $k$ from the original $100$ tuning runs. Tuning is performed on a validation set and the FID is reported on a separate test set.}
    \label{fig:tune_perf_two_scale}
\end{figure*}

\textbf{Tuning Study for Decoupled (Different) Learning Rates.} In Figure \ref{fig:tune_perf_two_scale}, we report the results of the tuning study when the base learning rates of $G$ and $D$ are tuned independently. The results conform with the conclusions we had from the same-rate setting: the scheduler outperforms its no-scheduler counterpart for most tuning budgets.

\begin{figure*}[htbp]
    \centering
    \includegraphics[scale=0.33]{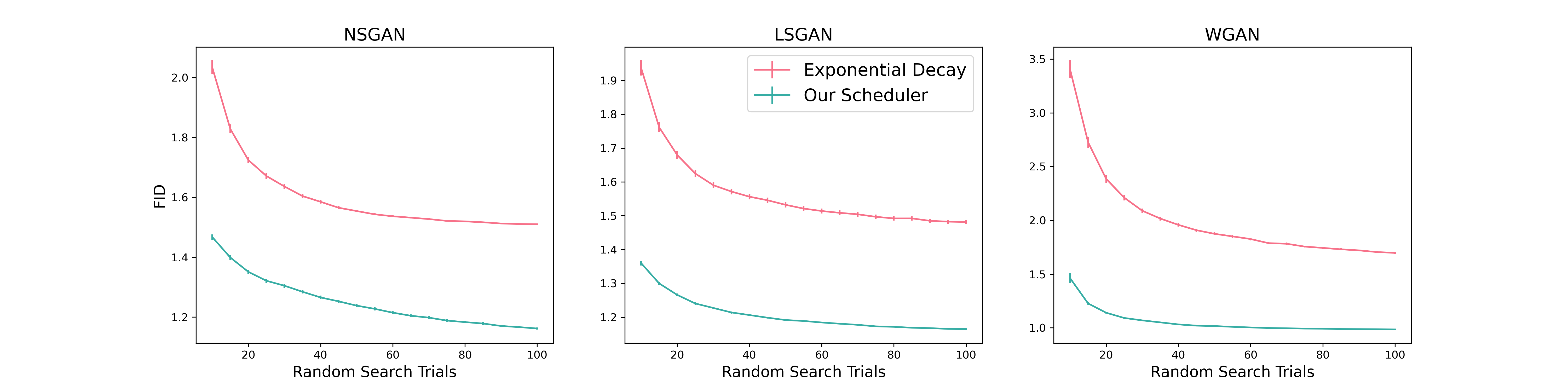}
    \caption{Exponential Decay vs. Our Scheduler. The exponential decay scheduler decays the base learning rate monotonically (i.e., does not depend on the current state of the network). The exponential decay factor is tuned.}
    \label{fig:exp_decay}
\end{figure*}

\textbf{Exponential Decay vs. Our Scheduler.} In Figure \ref{fig:exp_decay}, we compare our scheduler versus a (classical) exponential decay scheduler that decays the learning rate monotonically (i.e., does not depend on the loss of the network), on MNIST. We apply exponential decay at every step, in which the base learning rate is multiplied by $\rho^{s/T}$ where $s$ is the index of the current step and $T$ is the total number of steps. We tune both models (similar to the tuning experiment in the main paper), including the decay factor $\rho$ which we sample from a log-uniform distribution over the range $[10^{-4}, 0.1]$. 

\textbf{Scheduling Function: Exponential vs. Linear Interpolation.} Recall that we use exponential interpolation in the scheduling functions $f(x)$ and $h(x)$. Here we compare with linear interpolation, which is a natural alternative. For each interpolation method, we tune a GAN on MNIST, with the same architecture and tuning setup described in Section \ref{sec:experiments}. Using the best hyperparameters, we then train the GAN 100 times (using random seeds) and report the test FID. We report the results for both interpolation methods in Table \ref{table:linear_vs_exp}. The results indicate that exponential interpolation performs slightly better than linear interpolation for all the three GAN types considered.

\textbf{G's Optimality Gap.} While the scheduler is designed to control the learning rate (and consequently the gap) of $D$, we note that the scheduler also indirectly controls the optimality gap of $G$. Specifically, we define G's optimality gap as the absolute difference between G's training and ideal losses. G's ideal loss can be derived similar to that of $D$; e.g., for NSGAN it is $\log(2)$. In Table \ref{table:gen_gap}, we report G's optimality gap for GANs trained with and without the scheduler. The results indicate that the scheduler (which only controls D's LR) can significantly reduce G's optimality gap (by up to 60x).

\begin{table}[htbp]
\caption{Test FID of Exponential vs. Linear scheduling. We report the mean and standard error over 100 training runs (after tuning both).}
\label{table:linear_vs_exp}
      \centering
        \renewcommand{\arraystretch}{1.5}
        \scalebox{0.7}{\begin{tabular}{c|cc|}
            \cline{2-3}
                                     & \multicolumn{2}{c|}{FID (smaller is better)}                     \\ \cline{2-3} 
                                     & \multicolumn{1}{c|}{Exponential}       & Linear     \\ \hline
            \multicolumn{1}{|c|}{NS + Sched.} & \multicolumn{1}{l|}{\textbf{1.23} (0.02)} & 1.26 (0.04) \\
            \multicolumn{1}{|l|}{LS + Sched.} & \multicolumn{1}{l|}{\textbf{1.29} (0.02)} & 1.33 (0.03) \\
            \multicolumn{1}{|l|}{W + Sched.}  & \multicolumn{1}{l|}{\textbf{0.98} (0.02)} & 1.07 (0.02) \\ \hline
            \end{tabular}}
\end{table}

\begin{table}[htbp]
        \caption{{G's optimality gap (absolute difference between G's training and ideal losses) multiplied by $10^3$}. We report the mean and standard error over 100 training runs. Our scheduler, which only controls D, significantly reduces the optimality gap of G, compared to no scheduler. Asterisk (*)  means statistically significant based on a t-test at a level of 0.01.}
        \label{table:gen_gap}
      \centering
      \setlength{\tabcolsep}{6pt}
      \scalebox{0.7}{
        \begin{tabular}{|l|cccc|}
        \hline
        & \multicolumn{4}{c|}{Generator's Optimality Gap $\times 10^3$} \\
        GAN/Dataset  & MNIST  & Fashion  & CIFAR  & CelebA \\ \hline
        NS          & 90 (7) & 55 (4) & 308 (18) & 994 (26)\\
        NS + Sched. & \textbf{76} (6) & \textbf{48} (4) & \textbf{155} (6)*  & \textbf{358} (9)*\\ \hline
        LS          & \textbf{48} (4) & 29 (2) & 194 (0.01)  & 380 (0.007)\\ 
        LS + Sched. & 50 (4) & \textbf{22} (2) & \textbf{74} (1)*  & \textbf{145} (4)*\\ \hline
        W           & 138685 (6771) & 8065 (465) & 16883 (491)  & 7141 (520)\\ 
        W + Sched.  & \textbf{2320} (189)* & \textbf{5211} (155)* & \textbf{9597} (172)*  & \textbf{1612} (33)*\\ \hline
        \end{tabular}}
\end{table}

\section{Experimental Details} \label{appendix:experimental_details}
\textbf{Computing Setup:} We ran the experiments on a cluster equipped with P100 GPUs (we do not report the specs of the cluster for confidentiality). The tuning experiments took roughly $6$ GPU years. All models were implemented and trained using TensorFlow 2 \citep{tensorflow2015-whitepaper}, ran in GPU mode.

\subsection{GANs}
\textbf{Datasets and Processing: } For MNIST and Fashion MNIST, we use 50,000 examples for training, and 10,000 examples for each of the validation and test sets. For CIFAR, we use 40,000 examples for training, and 10,000 for each of the validation and test sets. For CelebA, we use the standard training set of 162,770 examples, and uniformly sample 10,000 examples for validation and testing from the standard validation/test sets, when computing FID. All pixels are rescaled to $[-1, 1]$, and for CelebA we resize all images to $32 \times 32$ (to reduce the memory requirements during training).

\textbf{FID and Inception Score: } FID and the Inception Score are computed using TF-GAN\footnote{https://github.com/tensorflow/gan} based on the Inception model, with the exception of  MNIST,  where TF-GAN uses a model trained on MNIST (with 99\% accuracy). For both measures, we use 10,000 generated images, and additionally for FID we use 10,000 real images from either the validation or test sets (depending on whether the validation or test FID is being computed).

\textbf{Training: } Validation FID is computed every 10 epochs for MNIST, 100 epochs for CIFAR-10 and Fashion MNIST, and 50 epochs for CelebA. Note that we compute FID less often for datasets other than MNIST because the FID  computation is expensive (each FID evaluation  can take more than 15 minutes on a GPU).

\textbf{Hyperparameter Ranges:} We denote a uniform distribution supported on $[a,b]$ by  $U(a,b)$ . Moreover, $L(a, b)$ denotes a log-uniform distribution: $x \sim L(a, b) \iff x \sim 10^{U(\log_{10}(a), \log_{10}(b))}$. The hyperparameters are sampled as follows: Learning rate from $L(10^{-5}, 10^{-3})$, $\beta_1$ (for Adam) from $U(0, 1)$, and WGAN clipping parameter from $L(10^{-3}, 1)$. Also, note that as discussed in the main paper, we tune the number of epochs by evaluating FID periodically during training.

\textbf{Architectures: } In all architectures, the generator is supplied with a $128$-dimensional noise vector, sampled from a standard normal distribution.

\textsl{MNIST and Fashion MNIST: } We use a  standard DCGAN architecture (taken from TensorFlow Core examples): 
\begin{itemize}
    \item Discriminator: Convolution ($64$ filters, $5 \times 5$ kernel, stride $2$, leaky ReLU, batchnorm, dropout 0.3) $\to$  Convolution ($128$ filters, $5 \times 5$ kernel, stride $2$, Leaky ReLU, batchnorm, dropout 0.3) $\to$ Flatten $\to$ Dense (1 unit). 
    \item Generator: Dense ($7 \times 7 \times 256)$, ReLU, batchnorm) $\to$ Reshape to $(7, 7, 256)$ $\to$ Up Convolution ($128$ filters, $5 \times 5$ kernel, stride $1$, ReLU, batchnorm) $\to$ Up Convolution ($64$ filters, $5 \times 5$ kernel, stride $2$, ReLU, batchnorm) $\to$ Up Convolution (1 filter, $5 \times 5$ kernel, stride $2$, Tanh).
\end{itemize}

\textsl{CIFAR and CelebA:} We use the standard   architecture for CIFAR in TF-GAN; with added  batchnorm and dropout (to conform with DCGAN).
\begin{itemize}
    \item Discriminator:  Convolution ($64$ filters, $5 \times 5$ kernel, stride $2$, leaky ReLU, batchnorm, dropout 0.3) $\to$  Convolution ($128$ filters, $5 \times 5$ kernel, stride $2$, Leaky ReLU, batchnorm, dropout 0.3) $\to$ $\to$  Convolution ($256$ filters, $5 \times 5$ kernel, stride $2$, Leaky ReLU, batchnorm, dropout 0.3) $\to$ Flatten $\to$ Dense (1 unit). 
    \item Generator: Dense ($4 \times 4 \times 256)$, ReLU, batchnorm) $\to$ Reshape to $(4, 4, 256)$ $\to$ Up Convolution ($128$ filters, $5 \times 5$ kernel, stride $2$, ReLU, batchnorm) $\to$ Up Convolution ($64$ filters, $4 \times 4$ kernel, stride $2$, ReLU, batchnorm) $\to$ Up Convolution (3 filters, $4 \times 4$ kernel, stride $2$, Tanh).
\end{itemize}

\subsection{DANN}

\textbf{Dataset and Processing: } We use 60,000 images from each of MNIST (labelled) and MNIST-M (unlabelled) during training. We use 5000 samples for each of the validation and test sets in MNIST-M. All pixels in the images are rescaled to $[-1,1]$.

\textbf{Hyperparameter Ranges:} The hyperparameters are sampled as follows: Learning rate for SGD from $L(10^{-4}, 10^{-2})$, $\lambda$ uniformly from $\{ 0.01, 0.1, 1 \}$, and $V^{*}$ from $U(0.5 \log(4), \log(4))$. As discussed in the main text, the number of epochs is tuned during each training run.

\textbf{Architecture: } We consider a simple CNN architecture similar to that in \citet{ganin2016domain} (with additional batchnorm and dropout to improve generalization). Below are the architecture details:
\begin{itemize}
    \item Feature Extractor: Convolution ($32$ filters, $5 \times 5$ kernel, stride $1$, ReLU, maxpooling, batchnorm) $\to$ Convolution ($48$ filters, $5 \times 5$ kernel, stride $1$, ReLU, maxpooling, batchnorm) $\to$ Flatten $\to$ Dropout (0.3).
    \item Label Predictor: Dense (100 units, ReLU) $\to$ Dense (100 units, ReLU) $\to$ Dense (10 units, Sigmoid).
    \item Discriminator: Dense (100 units, ReLU) $\to$ Dense (1 unit, Sigmoid).
\end{itemize}

\end{document}